\documentclass{article}

\PassOptionsToPackage{numbers, compress}{natbib}

\usepackage[preprint]{neurips_2025}

\usepackage[utf8]{inputenc} 
\usepackage[T1]{fontenc}    
\usepackage{url}            
\usepackage{booktabs}       
\usepackage[table,dvipsnames]{xcolor}
\usepackage{amsfonts}       
\usepackage{nicefrac}       
\usepackage{microtype}      
\usepackage{graphicx}
\usepackage[pagebackref, breaklinks, colorlinks]{hyperref}
\usepackage{wrapfig} 
\usepackage{multirow} 
\usepackage{subcaption}
\usepackage{makecell}
\usepackage{amssymb}
\usepackage{amsmath}
\usepackage{booktabs}
\usepackage[frozencache]{minted}
\usepackage[inline]{enumitem}
\usepackage[skip=1em]{caption}
\usepackage{wrapfig}
\usepackage{makecell}
\usepackage{vcell}
\usepackage{xspace}
\usepackage{adjustbox}
\usepackage{wrapfig}
\definecolor{brandeisblue}{rgb}{0.0, 0.44, 1.0}
\definecolor{SteelBlue}{HTML}{4682B4}
\definecolor{TealGreen}{HTML}{3CB371}
\definecolor{Amber}{HTML}{FFBF00}
\definecolor{Terracotta}{HTML}{D2691E}
\definecolor{cornellred}{rgb}{0.7, 0.11, 0.11}
\definecolor{cadmiumgreen}{rgb}{0.0, 0.42, 0.24}
\definecolor{aliceblue}{rgb}{0.91, 0.94, 0.97}
\definecolor{darkblue}{rgb}{0.83, 0.89, 0.97}
\definecolor{Red7}{rgb}{0.941, 0.243, 0.243}
\definecolor{Green7}{RGB}{55, 178, 77}
\definecolor{Blue9}{rgb}{0.098,0.3,0.9}

\usepackage{pifont}

\usepackage[dvipsnames]{xcolor}

\usepackage{titletoc} 
\definecolor{road}{RGB}{128,64,128}
\definecolor{sidewalk}{RGB}{232,35,244}
\definecolor{building}{RGB}{70,70,70}
\definecolor{wall}{RGB}{156,102,102}
\definecolor{fence}{RGB}{153,153,190}
\definecolor{pole}{RGB}{153,153,153}
\definecolor{trafficlight}{RGB}{30,170,250}
\definecolor{trafficsign}{RGB}{0,220,220}
\definecolor{vegetation}{RGB}{35,142,107}
\definecolor{terrain}{RGB}{152,251,152}
\definecolor{sky}{RGB}{180,130,70}
\definecolor{person}{RGB}{60,20,220}
\definecolor{rider}{RGB}{0,0,255}
\definecolor{car}{RGB}{142,0,0}
\definecolor{truck}{RGB}{70,0,0}
\definecolor{bus}{RGB}{100,60,0}
\definecolor{train}{RGB}{100,80,0}
\definecolor{motorcycle}{RGB}{230,0,0}
\definecolor{bicycle}{RGB}{32,11,119}
\definecolor{unlabeled}{RGB}{0,0,0}
\definecolor{CAR}{RGB}{0,0,142}
\definecolor{TRUCK}{RGB}{0,0,70}
\definecolor{PEDESTRIAN}{RGB}{182,1,14}
\definecolor{BIKE}{RGB}{119,11,32}
\definecolor{TERRAIN}{RGB}{152,251,152}
\definecolor{ROAD}{RGB}{128,64,128}
\definecolor{SIDEWALK}{RGB}{244,35,232}
\definecolor{SKY}{RGB}{70,130,180}
\definecolor{TRAFFICLIGHT}{RGB}{250,170,30}
\definecolor{FENCE}{RGB}{190,154,154}
\definecolor{TRAFFICSIGN}{RGB}{220,220,0}
\definecolor{LANELINE}{RGB}{255,255,255}
\definecolor{CROSSWALK}{RGB}{55,176,189}
\definecolor{BUS}{RGB}{0,60,100}
\definecolor{BUILDING}{RGB}{70,70,70}
\definecolor{WALL}{RGB}{102,102,156}
\definecolor{POLE}{RGB}{153,153,153}
\definecolor{VEGETATION}{RGB}{107,142,35}

\usepackage[dvipsnames]{xcolor}
\def\onedot{\ifx\@let@token.\else.\null\fi\xspace}
\def\eg{\emph{e.g}\onedot}

\newcommand{\method}{{Dreamland}\xspace}

\title{\method: Controllable World Creation with Simulator and Generative Models}

\author{%
  Sicheng Mo\thanks{Equal Contribution} \quad
  Ziyang Leng\footnote[1]{} \quad
  Leon Liu\quad
  Weizhen Wang\quad
  Honglin He\quad
  Bolei Zhou\\
  [2mm]
  University of California, Los Angeles 
}

\begin{document}

\maketitle

\begin{figure}[h]
    \centering
    \vspace{-2em}
    \includegraphics[width=1.0\linewidth]{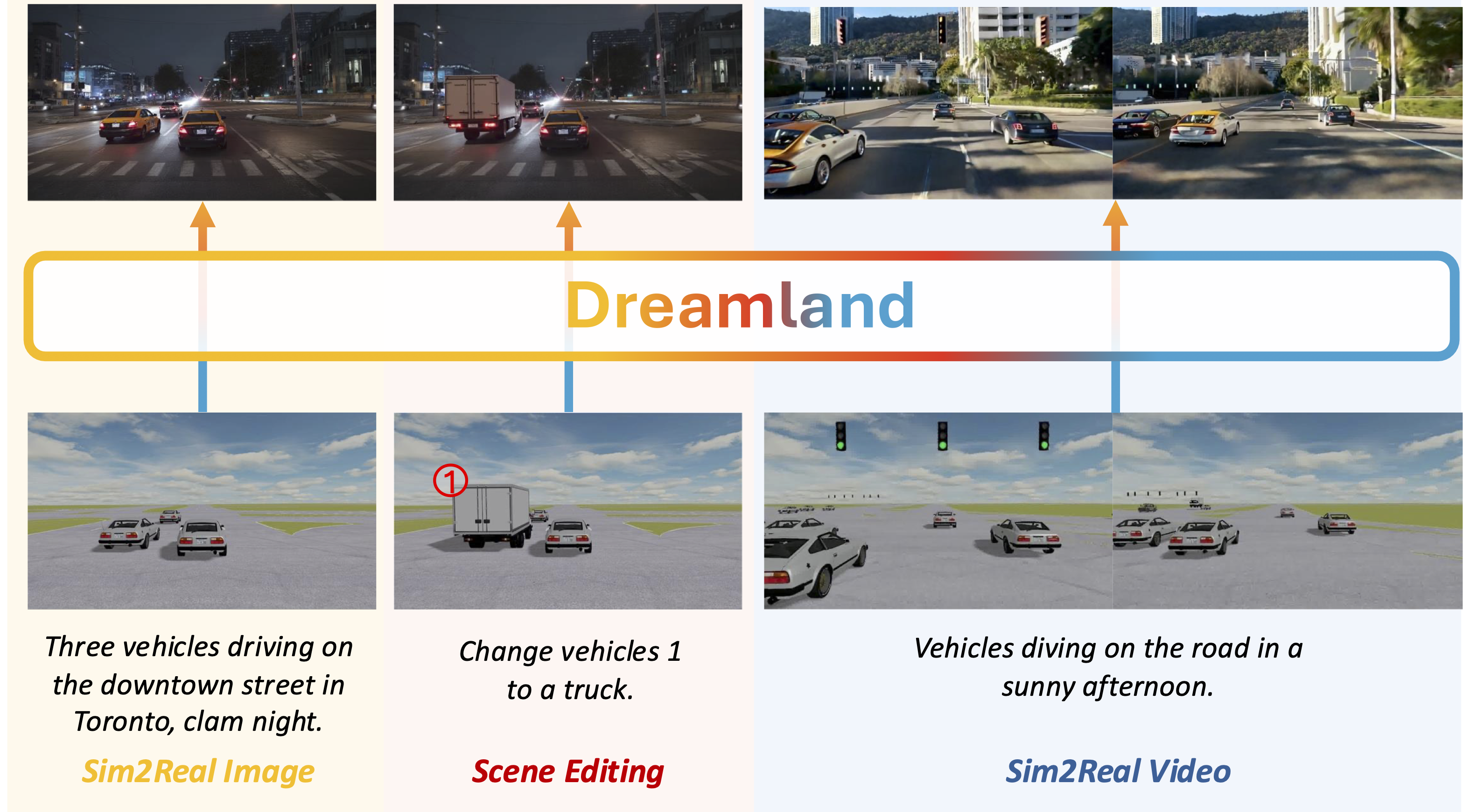}
    \caption{
    \textbf{Controllable world creation with Dreamland}. It combines a simulator for physically grounded scene generation and a large-scale pretrained generative model for creating a realistic visual world following user-provided text prompts. 
    }
    \label{fig:teaser}
\end{figure}

\begin{abstract}
Large-scale video generative models can synthesize diverse and realistic visual content for dynamic world creation, but they often lack element-wise controllability, hindering their use in editing scenes and training embodied AI agents.
We propose \method, a hybrid world generation framework combining the granular control of a physics-based simulator and the photorealistic content output of large-scale pretrained generative models. In particular, we design a layered world abstraction that encodes both pixel-level and object-level semantics and geometry as an intermediate representation to bridge the simulator and the generative model. This approach enhances controllability, minimizes adaptation cost through early alignment with real-world distributions, and supports off-the-shelf use of existing and future pretrained generative models. 
We further construct a D3Sim dataset to facilitate the training and evaluation of hybrid generation pipelines. Experiments demonstrate that \method outperforms existing baselines with $50.8\%$ improved image quality, $17.9\%$ stronger controllability, and has great potential to enhance embodied agent training. Code and data will be made available at \url{https://metadriverse.github.io/dreamland/}. 

\end{abstract}

\section{Introduction}
\label{sec: intro}
Large-scale pre-training with scalable model architectures has significantly advanced generative modeling in recent years. This progress has led to a surge of foundation models across various modalities, including language~\cite{llama2,llama3,gemma,zhu2023minigpt,liu2023llava,liu2023improvedllava}, image~\cite{rombach2022ldm,podell2023sdxl,esser2024sd3,nichol2021glide,ramesh2021dalle,ramesh2022dalle2}, and video generation~\cite{openaiSoraCreating,zheng2024opensora,lin2024opensora-plan,ma2024latte}. 
These foundation models capture structured information about entities, their interactions, and temporal dynamics—often referred to as the "world knowledge".
As such, they hold great promise for providing synthetic data to train embodied AI agents beyond just generating visual content. By providing interactive feedback, they can potentially replace human supervision or physical environments, enabling more efficient and scalable agent learning.

Despite the capabilities of current world models being exciting, they often lack the fine-grained control required for agent learning, e.g., autonomous driving. 
For instance, training autonomous vehicles requires simulating complex scenarios. These include specific vehicle maneuvers like lane changes in dense traffic or responses to stop signs. 
Current generative world models often struggle to offer object-level controllability, thereby constraining their efficacy in situations where precise scene layout and object configuration are critical.

Recognizing these challenges, a hybrid approach that combines physical simulators and data-driven generative models has emerged as a promising solution. 
In such pipelines, the simulators first provide the accurate physical and spatial information, such as vehicle dynamics, traffic rules, and environmental function zones. Then, image or video generative models synthesize realistic visual content following these conditions.
Despite these improved pipelines, current hybrid methods struggle to balance simulator fidelity with generative freedom.
Some methods directly render scenes without any hallucination~\cite{yu2024lucidsim,li2025simworld}, which necessitates high-fidelity simulators and compromises their generalizability and scalability for diverse world creation. On the other hand, other work~\cite{zhou2024simgen} fully utilizes generative models for scene re-rendering without stringent constraints, which consequently sacrifices fine-grained simulator control.

To balance the controllability from simulators and the creative freedom of generative models with rich visual details in a synergetic way, we present \method, bridging simulator and generative models through Layered World Abstraction (\textbf{LWA}). LWA records structured meta information that contains pixel-level information like depth and RGB value, and object-level information like object categories.
Instead of directly generating the scene from the initial simulator layered world abstraction~(\textbf{Sim-LWA}), \method first augments it to real-world layered world abstraction~(\textbf{Real-LWA}) that better aligns with real-world visual distributions. 
During inference, \method follows user instructions to divide the scene into preserved and editable regions. Editing models could help diversify the scene structure within the editable region while keeping the preserved region untouched.
Therefore, by strictly following conditions in Real-LWA during visual re-rendering, the final generated scenes can achieve both high-quality appearance and precise object control that adheres to the simulator and user instructions.

\method design offers several strengths. First, our pipeline largely enhances controllability and flexibility over the hybrid approach, outperforming the previous state-of-the-art~\cite{zhou2024simgen} by $52.3\%$ and $17.9\%$ in image quality and controllability. Notably, \method also demonstrates benefits in downstream agent training, boosting visual question answering performance on the real-world test set by $3.9$ absolute accuracy. 
Second, since the Real-LWA is already aligned with real-world distributions, it can be seamlessly integrated with future, more potent conditional-generation models without introducing significant adaptation costs.
Finally, \method pipeline is generalizable and scalable, thus unlocking various applications based on simulators, including video generation and scene editing, as illustrated in Figure~\ref{fig:teaser}.
We also construct a large-scale dataset for training and evaluating such hybrid pipelines.

We summarize our key contributions as follows:(1) \textbf{\method}~ - a hybrid generation pipeline that connects a physics-based simulator and generative models with a layered world abstraction (\textbf{LWA}) to achieve controllable and configurable world creation.

(2) Our approach demonstrates superior results in the scene generation task, exceeding previous state-of-the-art by $52.8\%$ image quality and $17.9\%$ controllability. Also, we demonstrate how \method improves the adaptation of embodied agents to the real world.
(3) We construct a dataset called \textbf{D3Sim} (\textbf{D}iverse \textbf{D}riving Scenario in Real Worl\textbf{D} and \textbf{Sim}ulation) for training and benchmarking hybrid generation pipelines that combine simulators and generative models.

\begin{figure}[h]
    \centering
    \vspace{-2em}
    \includegraphics[width=1.0\linewidth]{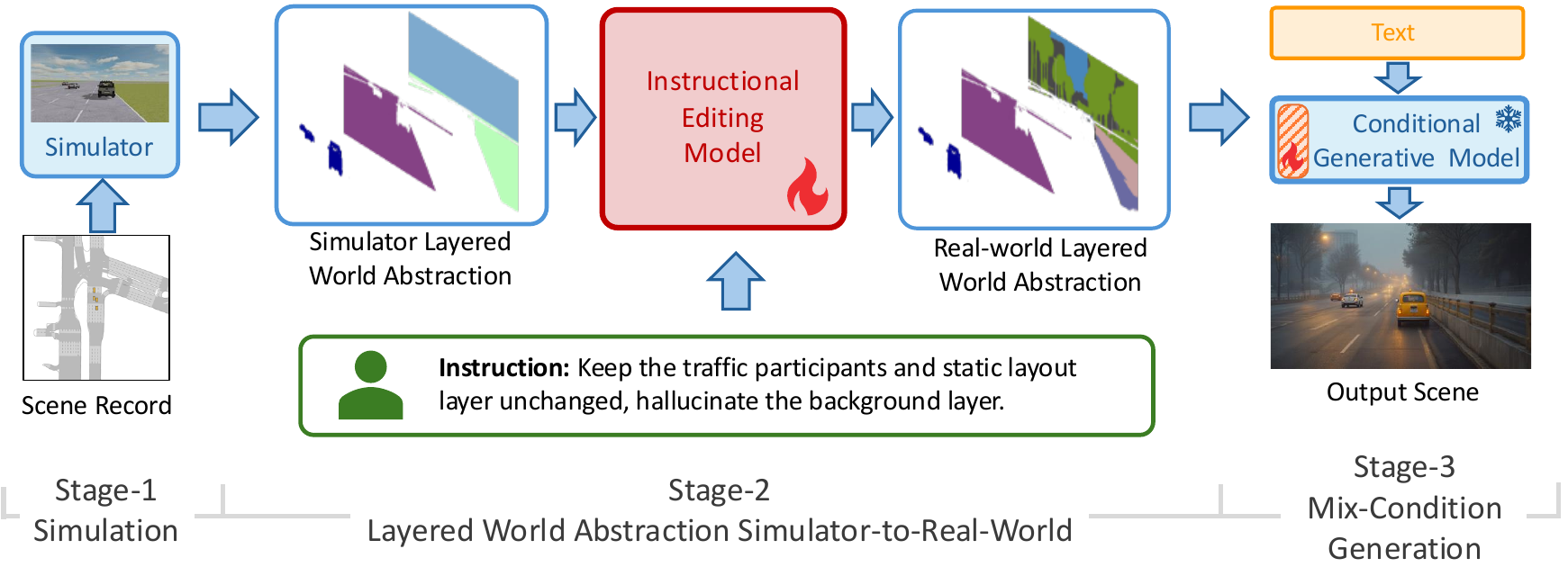}
    \vspace{-2.0em}
    \caption{
    \textbf{
    Illustration of the Dreamland}. It is a three-stage pipeline including: (1) Simulation, (2) Layered World Abstraction Simulator-to-Real-World, (3) Mix-Condition Generation. 
}
    \vspace{-1.5em}
    \label{fig:framework}
\end{figure}

\section{Related Work}
\label{sec: related works}

\vspace{-0.5em}
\smallskip
\noindent
\textbf{Controllable Visual Generative Models}.
Large-scale foundational generative models have been developed rapidly in recent years for image generation~\cite{rombach2022ldm,podell2023sdxl,esser2024sd3,nichol2021glide,ramesh2021dalle,ramesh2022dalle2}, video generation~\cite{openaiSoraCreating,zheng2024opensora,lin2024opensora-plan,ma2024latte}, and multimodal generation~\cite{zhou2024transfusion,xie2024show,shi2024llamafusion, Emu2, emu3, mo2025xfusion}. While those foundation models usually rely solely on text prompts, various approaches have been proposed to add additional controllability to the generation process, including structure control~\cite{zhang2023controlNet,li2023gligen,mou2023t2i,ju2023humansd,mo2023freecontrol,tumanyan2023plug-and-play,lin2024ctrl-x}, subject control~\cite{ruiz2023dreambooth,ye2023ipadapter,avrahami2023break-a-scene,mokady2023null}, and adding image editing capability~\cite{brooks2023instructpix2pix,mao2025ace++}. Meanwhile, another line of work addresses the more challenging multi-condition generation problem. Specifically, UniControl~\cite{zhao2023unicontrol} and UniControlNet~\cite{zhao2023unicontrolnet} propose unified adapters to process various conditioning signals. 
Distinguished from the above works that could obtain high-quality paired data with pre-trained detection models, the simulator and real-world paired data are expensive to annotate. Therefore, \method focuses on adapting from existing conditional generation models with low cost while preserving their rich world knowledge.

\smallskip
\noindent
\textbf{Generative Model for Autonomous Driving}. Generative models have largely advanced autonomous driving research in recent years. One line of work~\cite{bevgen,yang2023bevcontrol,gao2024magbeic_drive_v1,gao2024magicdrive_v2,wang2023drivedreamer,li2023drivingdiffusion} trains generative models, including GANs and Diffusion Models, to generate multi-view images from a driving scene graph, from bird-eye-view to HD maps. Specifically, Panacea~\cite{wen2024panacea}, InfiniteCube~\cite{lu2024infinicube}, MagicDrive~\cite{gao2024magbeic_drive_v1}, and MagicDrive-V2~\cite{gao2024magicdrive_v2} build on pre-trained video diffusion models to add temporal consistency into the generated scene. Another line of work~\cite{zheng2024genad,gao2024vista,hu2023gaia1generativeworldmodel,kim2021drivegan} takes historical information as the condition signal to generate future driving scenes, but they often lack control over the generated scenes. \method differs from previous literature in its hybrid approach, allowing fine-grained control over each driving vehicle, such as specifying their individual paths, actions, and interactions.

\smallskip
\noindent
\textbf{Synthetic Data for Embodied AI Training}.
The significant visual domain gap between simulators and the real world data has historically presented a challenge in training agents in simulation. Due to recent advances in generative and reconstruction techniques, this problem has been partially addressed by integrating simulators, vision generative models, and 3D reconstruction methods. For example, Vid2Sim~\cite{xie2025vid2sim} and VR-Robo~\cite{zhu2025vr-robo} employ Gaussian-Splatting techniques to convert a real scene to an interactive digital twin, and trained agents can zero-shot deploy to the real world. On the other hand, Lucidsim~\cite{yu2024lucidsim} trains a robot dog to achieve great adaptability using a simulator with a pre-trained depth-conditioned diffusion model. Compared to previous literature, the \method pipeline ensures granular controllability while being applicable to various generative models.

\vspace{-0.5em}
\section{Method}
\label{sec:method}
\vspace{-0.5em}
\method enables controllable driving scene creation via three key stages shown in Figure~\ref{fig:framework}: 
(i) \textit{Stage-1 Simulation}: scene construction with physics-based simulator. 
(ii) \textit{Stage-2 LWA-Sim2Real}: transferring the Sim-LWA from simulation to Real-LWA with an instructional editing model and user instructions. 
(iii) \textit{Stage-3 Mixed-Condition Generation}: rendering an aesthetic and realistic scene with a large-scale pretrained image or video generation model.
The pipeline starts with the simulator rendering a scene according to the scene record, where the agents' motion is derived from trajectory replay or trained policy. The Sim-LWA captured from the simulator is then refined and rendered into realistic frames.
We first introduce the layered world abstraction in \S \ref{sec:mlwp}. We then talk about the design of Stage-2 in \S \ref{sec:cond_trans}, and Stage-3 in \S \ref{sec:model_adapt}. Lastly, we describe our training scheme and implementation details in  \S \ref{sec:train_sche}.

\subsection{Layered World Abstraction}
\label{sec:mlwp}
We define Layered World Abstraction (LWA) as an intermediate representation to align simulators and generative models in the pixel space. It enables fine-grained control over the generation process. LWA composes a scene from multiple world layers, where each layer corresponds to different classes of objects or regions. The representation is structured as 
\vspace{-2mm}
{\small
\begin{equation}
    \mathcal{W} = \bigcup_{i=0}^N \mathcal{L}_i \odot \mathcal{V}_i, 
    \label{eq:wr}
\end{equation}
}
where $\mathcal{L}_i\in \mathbb{R}^{H \times W \times D}$ denotes the $i^{\text{th}}$ world layers, and $\mathcal{V}_i \in \mathbb{R}^{H \times W}$ is the corresponding visibility mask. Each world layer encodes both pixel-level and object-level geometry and semantics through $K$ modalities of condition. Thus, the $i^{\text{th}}$ layer is formulated as 
{\small
\begin{equation}
    \mathcal{L}_i = \left\{ \mathbf{c}_j \in \mathbb{R}^{H\times W\times C_j} \middle| j=0,1,\ldots, K; \sum_{j=0}^K C_j = D\right\},
\end{equation}
}where $\mathbf{c}_j$ denotes the $j^{\text{th}}$ condition with $C_j$ channels. This flexible design allows for customizable, pixel-level control over both the object-of-interest and the region-of-interest. 

Given a driving frame rendered by the simulator following the scene record, we decompose it into Sim-LWA according to the pixel semantics. Sim-LWA contains accurate traffic participants and layout, which is physically grounded by the simulator and serves well for layers requiring precise control. However, due to the limited assets and lack of background details, those corresponding layers hold a disparity from real-world distribution and hinder the realistic scene generation.

\subsection{Stage-2: LWA-Sim2Real}
\label{sec:cond_trans}
The LWA-Sim2Real refinement process involves refining the Sim-LWA to real-world distributions. In the \method pipeline, we compose our LWA from three world layers: a traffic participants layer $\mathcal{L}^d$ that contains dynamic objects (\emph{e.g.}, vehicles, pedestrians, cyclists), a map layout layer $\mathcal{L}^l$ that defines a static layout (\emph{e.g.}, roads, crosswalk, intersections), and a background layer $\mathcal{L}^b$ that covers static objects and background regions (\emph{e.g.}, buildings,  vegetation, sky). The first two layers, $\mathcal{L}^d$ and $\mathcal{L}^l$, are derived from the simulator and preserved to provide precise control, while the last layer $\mathcal{L}^b$ is editable and refined by an editing model as described below.

Given preserved world layers and their masks from simulators $\{\mathcal{L}^d, \mathcal{L}^l \},\{\mathcal{V}^d, \mathcal{V}^l \}$, we employ an instructional editing model $\epsilon_e$ to selectively refine layer $\mathcal{L}^b$ according to the provided text instruction $c$. This process can be formulated as,
{\small
\begin{equation}
    \mathcal{L}^b = \epsilon_e \left(\hat{\mathcal{W}}, \mathcal{V}^b, c \right).
    \label{eq:lb}
\end{equation}
} The $\hat{\mathcal{W}} = \left\{\mathcal{L}^d \odot \mathcal{V}^d \cup \mathcal{L}^l \odot \mathcal{V}^l \right\}$ denotes the preserved LWA from the simulator, and the $\mathcal{V}^b = \Omega \setminus (\mathcal{V}^d \cup \mathcal{V}^l)$ is the editing mask given the pixel domain $\Omega$. 

This process transfers the Sim-LWA to real-world distribution while preserving the grounding information from the simulator. It bridges the domain gap between the general simulators and generation models, alleviating the requirements for high-fidelity simulation rendering and minimizing the adaptation costs for generation models.

\subsection{Stage-3: Mixed-Condition Generation}
\label{sec:model_adapt}

In this stage, we utilize a pre-trained conditional generative model to generate realistic views according to our refined Real-LWA. As the representation is already aligned with real-world distribution, minimal adaptation is required to employ a pre-trained model on it, thus preserving its world knowledge. For pre-trained conditional world models that cover the condition modalities in our LWA, we split the channel dimension of LWA according to condition modalities and serve as the respective condition inputs.

However, some of the pre-trained conditional world models are trained on specific condition modalities that do not fully cover the modalities within our LWA. Therefore, we have two choices to further minimize the adaptation cost while preserving the pre-trained world knowledge: (1) Extracting only specific condition modalities, e.g., depth or segmentation maps, from LWA. (2) Adapting the pre-trained conditional world model to our LWA by updating a small number of parameters. We empirically found that the second approach yields better controllability thus adopted it in our design.

Given the Real-LWA, which contains $K$ conditions, we first encode them into the latent space with encoder $\mathcal{E}$, and concatenate them along the channel dimension. Then, we use a linear layer $\xi$ to project the concatenated condition latent into the same dimension as the noise latent $\mathbf{x}$ of the world model. Finally, we add the projected condition latent with the noise latent to incorporate structural guidance. The whole process can be formulated as
{\small
\vspace{-0.2em}
\begin{equation}
    \mathbf{x'} = \mathbf{x} + \xi\left( \bigoplus_{j=0}^K \mathcal{E}(\mathbf{c}_j)  \right).
    \label{eq:xprime}
\end{equation}
}

The output noise latent $\mathbf{x'}$ with structural information serves as the input to the diffusion transformer blocks of the world model. Thereby, we achieve mixed-condition generation by fine-tuning the projection layer $\xi$, rendering a realistic scene from the refined LWA.

\subsection{Training Scheme and Implementation Details}
\label{sec:train_sche}

We train our \method pipeline in two steps. The first training step corresponds to the LWA-Sim2Real stage, where it enables the instructional editing model to learn Sim2Real refining of LWA. The second step is for the mixed-condition generation stage, which involves fine-tuning a conditional generation model to take structural control from the Real-LWA. We design our instructional editing model following ACE++~\cite{mao2025ace++}, training on our image dataset for 4K iterations with a batch size of 128. To align with the pre-trained editing model, we expand the world representation into a single image by concatenating the condition maps vertically. The loss for LWA Sim2Real transfer follows
\vspace{-0.25em}
\begin{equation}
    L_\text{Sim2Real} =  \mathbb{E}_{\mathcal{L}^b_0}\left[\| \mathcal{L}^b - \mathcal{L}^b_0 \|^2_2\right],
\end{equation}
where $\mathcal{L}^b$ is the refined layer defined in Eq.~\ref{eq:lb} and $\mathcal{L}^b_0$ denotes the corresponding world layer in real-world distribution.
We employ Flux Depth ~\cite{flux2024}, an image generation model $\epsilon_{\theta}$ with structure control based on depth maps, for our second step and adapt it to our LWA using the loss formulated as
\begin{equation}
    L_\text{adapt} = \mathbb{E}_{\mathbf{x}_0}\left[\| \epsilon_{\theta} (\mathbf{x'}) - \mathbf{x}_0 \|^2_2\right], 
\end{equation}

where the $\mathbf{x'}$ is from Eq.~\ref{eq:xprime} and $\mathbf{x}_0$ is the latent of the realistic view. To balance the inference cost and visual quality, we set the default resolution(width) of our second and third-stage models to 512 and 1024, respectively. Additional details are provided in Appendix~\ref{sec:suppl_imple_details}.

\section{Data curation} 
\label{sec: data curation}
We curate a large-scale dataset called D3Sim (\textbf{D}iverse \textbf{D}riving Scenario in Real Worl\textbf{D} and \textbf{Sim}ulation) that contains diverse driving scenarios in the real world and the simulation. It provides realistic perspective views and high-quality condition data to facilitate the Sim2Real transfer. Previous digital twin driving datasets (e.g., DIVA-real \cite{zhou2024simgen}) comprise limited samples with the image mismatch problem, which is suboptimal for training our editing model. Thus, we precisely aligned our digital twin driving scenes for LWA-Sim2Real transfer. Based on the nuPlan dataset \cite{nuplan}, we obtain realistic conditions \emph{(Real Conditions)} using pre-trained models. We then utilize ScenarioNet \cite{li2023scenarionet} to construct the corresponding digital twin simulation scene in MetaDrive simulator \cite{li2021metadrive}. By replaying the ego vehicle trajectory, we obtain conditions in the simulation domain \emph{(Sim Conditions)}. We organize the constructed digital twins into two dataset variants for the \method pipeline. Details of the data curation pipeline, as well as the video dataset used in the \method-Video pipeline, are provided in the Appendix~\ref{sec:suppl_data}.

\smallskip
\noindent
\textbf{Training Dataset}.
This dataset contains digital twin driving scenarios that capture the real-world distribution, enabling the second-stage model to learn Sim2Real transfer. We construct the dataset with paired Sim and Real conditions, along with realistic perspective views. We split the conditions into three world layers according to the semantics and construct the LWA accordingly. This results in high-quality digital twin training data from approximately 1,800 scenarios with around 60,000 samples at a 2 Hz sample rate. We use this dataset to train our \method pipeline.

\begin{wrapfigure}{r}{0.35\textwidth}
\begin{minipage}[b]{0.35\textwidth}
\centering
\includegraphics[width=\textwidth]{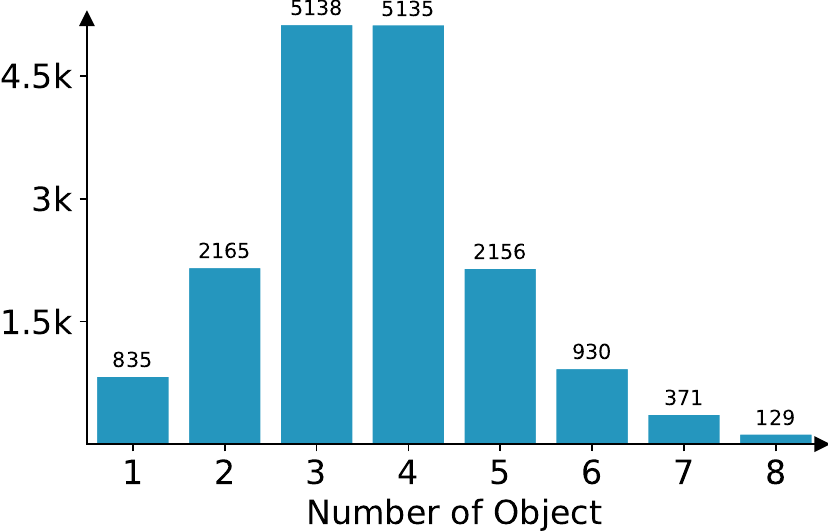}
\vspace{-1.5em}
\caption{Object distribution of the validation dataset.}
\label{fig:obj_distri}

\end{minipage}
\end{wrapfigure}

\smallskip
\noindent
\textbf{Validation Dataset}. The validation dataset is used to evaluate both the visual quality and controllability of the generation pipeline. While the digital twin scenarios provide realistic views and distribution of the real-world driving, the complex scene compositions, including partially occluded or overlapping objects, affect the accurate evaluation of controllability. Therefore, we derived this validation dataset from real-world scenarios by selectively modifying the traffic participants, eliminating ambiguous object placements, while keeping the scene layout unchanged. Diverse text prompts involving more than 30 cities, 20 weather conditions, different times of day, and various street types are randomly generated for each sample. This yields diverse and clean validation data from more than 600 scenarios with 16,000 samples at a 2 Hz sample rate, with object distribution shown in Figure.~\ref{fig:obj_distri}.

\begin{figure}[]
    \centering
    \vspace{-0.5em}
    \includegraphics[width=1.0\linewidth]{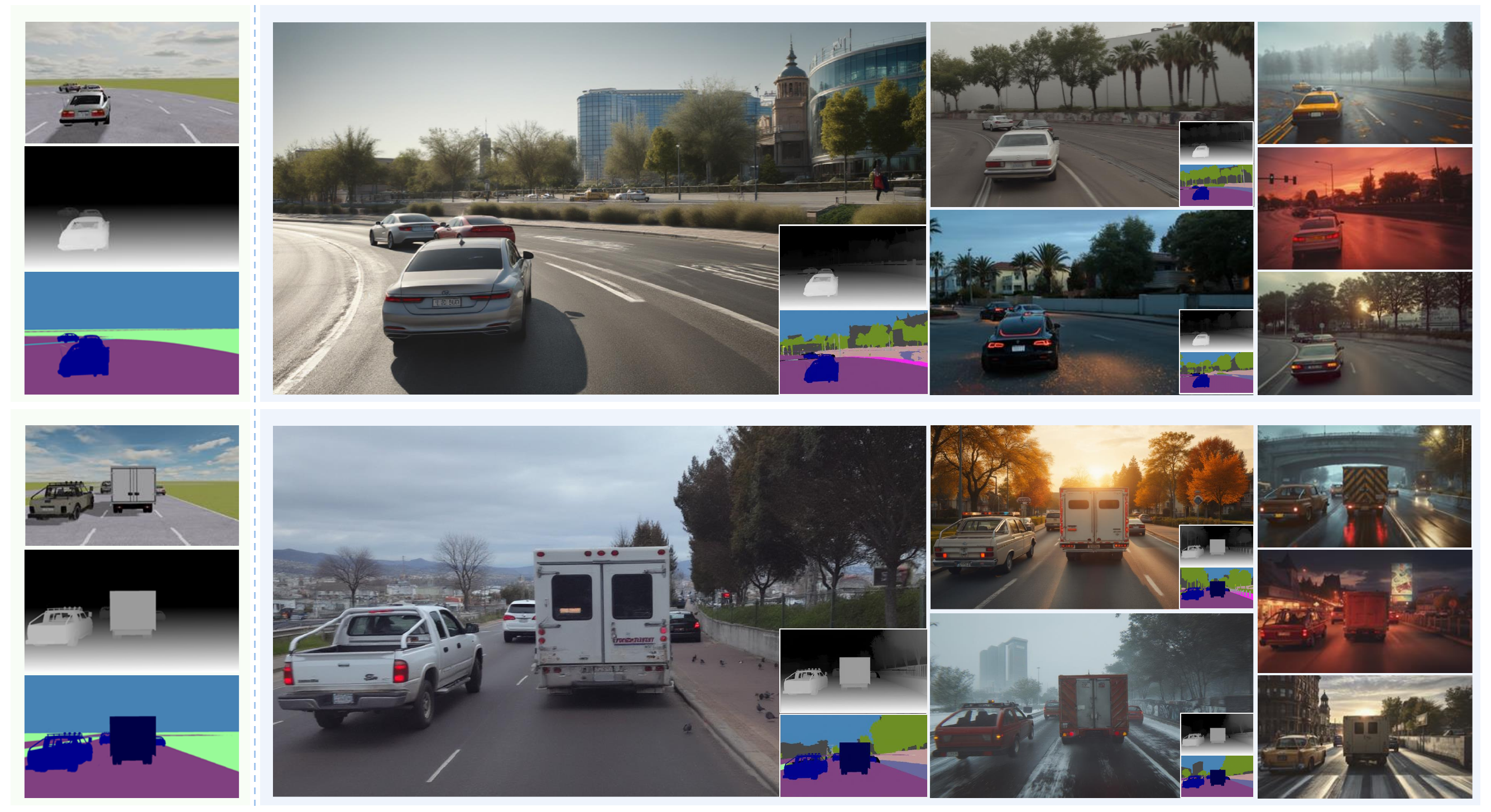} 
    \vspace{-1.7em}
    \caption{
    \textbf{Rendering diverse appearances align to the simulator’s conditions}. \method can simulate the driving scene and render it into high-quality scenes with generative models.
    }
    \vspace{-1em}
    \label{fig: qual of dreamland}
\end{figure}

\section{Experiments}
\label{sec: experiments}
Extensive qualitative and quantitative results demonstrate the proposed pipeline's generation performance (Sec.~\ref{subsection: simulator re-rendering}). We also present experiments on flexible control using \method, an extension to \method-Video for continuous simulator re-rendering (Sec.~\ref{subsection: video pipeline}), an ablation study of our key designs (Sec.~\ref{section: ablation}), and an application to a downstream agent learning task (Sec.~\ref{section:agent training}).

\smallskip
\noindent
\textbf{Evaluation Metrics.}
There are two aspects regarding the evaluation of the simulator-controlled world creation pipeline: the fidelity of the re-rendered scene to the original simulator output, and the overall visual quality of these visual re-renderings. We render 16K scenes from our D3Sim validation dataset and report the FID to DIVA-Real Dataset constructed in SimGen~\cite{zhou2024simgen}. We also evaluate depth si-RMSE and segmentation mIoU following previous works~\cite{zhang2023controlNet,nvidia2025cosmos} for rendered scenes.

\subsection{Simulator Conditioned Generation}
\label{subsection: simulator re-rendering}

\smallskip
\noindent
\textbf{Qualitative Results.} As shown in Figure~\ref{fig: qual of dreamland}, \method is able to re-render a scene into diverse high-quality scenes closely aligned with the spatial scene layout in the simulator but differ in weather, location, and light condition following user-provided text prompts. Leveraging the robust capabilities of pre-trained Text-to-Image diffusion models, Dreamland is capable of synthesizing realistic driving scenes with a resolution of up to 1024 $\times$ 576 pixels.

\smallskip
\noindent
\textbf{Baselines.} Our main baselines include generative models for autonomous driving that take scene layout conditions. 
SimGen~\cite{zhou2024simgen} first obtains front-view observation from the MetaDrive simulator~\cite{li2021metadrive}, then trains cascade diffusion models to render high-quality images. Alongside our main approach, we create variations of \method by extracting depth condition from our LWA and directly employing pretrained depth-conditioned diffusion models for as third-stage models, notated as \method (\textit{Frozen \{Model\}}).

\smallskip
\noindent
\textbf{Comparison with Baselines.}

\begin{wraptable}{R}{0.55\textwidth}
    \centering
    \vspace{-1em}
    \begin{minipage}{0.55\textwidth}
    \begin{scriptsize}
    \resizebox{\linewidth}{!}{
        \begin{tabular}{lccc}
\toprule[1.5pt]
                        & \multicolumn{1}{c}{Image Quality} & \multicolumn{2}{c}{Controllability}  \\
\cmidrule(r){2-2} \cmidrule(r){3-4}
Method                  & FID~$\downarrow$        & si-RMSE~$\downarrow$ & mIoU~$\uparrow$   \\
\midrule[0.5pt] 

SimGen                  & 106.02                     & 0.787      & 0.752             \\
\addlinespace
\textbf{Dreamland} (\textit{Frozen SDXL}) & 84.48                           & 0.744      & 0.698             \\
\textbf{Dreamland} (\textit{Frozen SD3}) & 63.74                           & 0.673      & 0.747             \\
\textbf{Dreamland} (\textit{Frozen Flux}) & 48.48                       & 0.639      & 0.785             \\
\rowcolor{gray!10}\textbf{Dreamland} (\textit{Flux})        & 50.58                       & 0.646      & 0.791            
\\
 \bottomrule[1.5pt]
\end{tabular}
    }
    \end{scriptsize}
    \vspace{-0.5em}
    \caption{\textbf{Comparison on image quality and controllability}. \method is strong and scalable.}
    \vspace{-0.2em}
\label{tab:quant_methods}
    \end{minipage}
\end{wraptable}

\begin{figure}[]
    \centering
    \vspace{-2.0em}
    \includegraphics[width=1.0\linewidth]{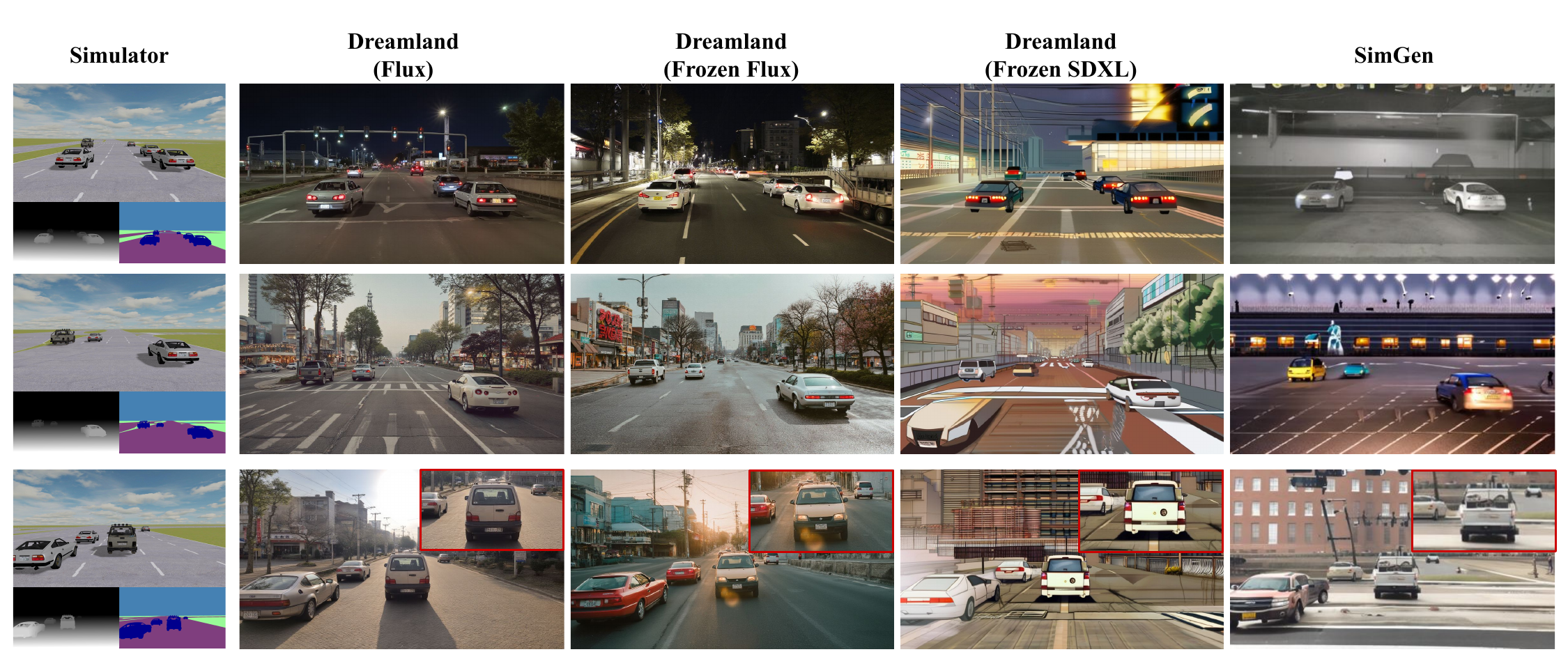} 
    \vspace{-2.0em}
    \caption{
    \textbf{Qualitative comparison} with baseline methods.
    \method with fine-tuned Flux matches the visual fidelity of the frozen Flux model while improving adherence to simulator conditions.
    \vspace{-1.5em}
    }
    \label{fig:qual compare baselines}
\end{figure}

Figure \ref{fig:qual compare baselines} and Table \ref{tab:quant_methods} compare our methods against several baselines. We observe that \method vastly outperforms SimGen with 52.3$\%$ lower FID and 17.9$\%$ better si-RMSE. Meanwhile, the Dreamland pipeline exhibits superior scalability: it benefits from plugging in progressively stronger pretrained models for Stage-3 (SDXL $\rightarrow$ SD3 $\rightarrow$ Flux), yielding marked gains in image quality and controllability. Notably, Dreamland fine-tuned on Flux matches the visual fidelity of the frozen Flux model while slightly improving adherence to simulator conditions, highlighting the world-knowledge preservation inherent in our \method pipeline.

\vspace{-1em}
\subsection{Extension of \method.}

\begin{figure}[]
    \centering
    \includegraphics[width=0.85\linewidth]{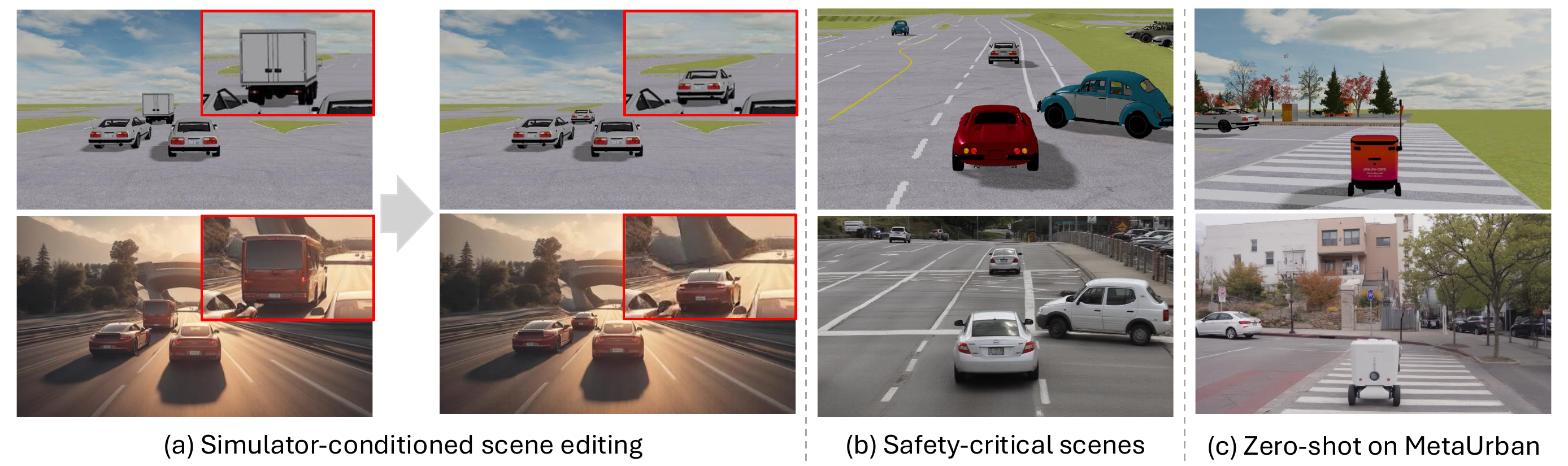} 
    \vspace{-1.0em}
    \caption{
    \textbf{Extension of \method}. \method pipeline is generalized to various downstream tasks.
    \vspace{-1.5em}
    }
    \label{fig: qual tasks}
\end{figure}

\noindent
\textbf{Simulator Editing.}
Figure~\ref{fig: qual tasks} (a) shows that \method supports a new application that edits a generated scene by adjusting the corresponding source scene. Building upon our current pipeline, we employ a pretrained Flux-fill model to remove the truck and add a speedy car on the road. By updating the LWA, we could provide additional editing mask support, such as a pretrained model without additional adaptation cost, showcasing the flexibility of our \method pipeline. 

\noindent
\textbf{Safety-critical Generation.} Figure~\ref{fig: qual tasks} (b) shows that by designing the driving scenario with the simulator, \method could generate safety-critical scenes that are dangerous to collect in real world, revealing the great potential of applying \method for agent learning.

\noindent
\textbf{Extension to Additional Simulator.} With the simple and generalized pipeline design, our LWA naively supports various simulator types. Figure~\ref{fig: qual tasks} (c) shows \method's zero-shot capability on the MetaUrban~\cite{wu2025metaurban} simulator for mobility agent learning.

\noindent
\textbf{Extension to Video Re-Rendering.}
\label{subsection: video pipeline}
Benefit from the simple pipeline design and low adaptation cost of \method,  it extends naturally to video generation without any specialized architectural changes. In Stage-2, we fine-tune the Cosmos-Predict1-7B~\cite{nvidia2025cosmos} Text-to-Video diffusion model into an instructional video-editing model. The training uses our D3Sim video dataset for 10K iterations with a batch size of 8. Following standard practice in image editing models, we initialize from the pretrained generative model and concatenate the extra conditioning latents with the noise latent along the channel dimension.

\begin{figure}[]
    \centering
    \vspace{-0.5em}
    \includegraphics[width=0.98\linewidth]{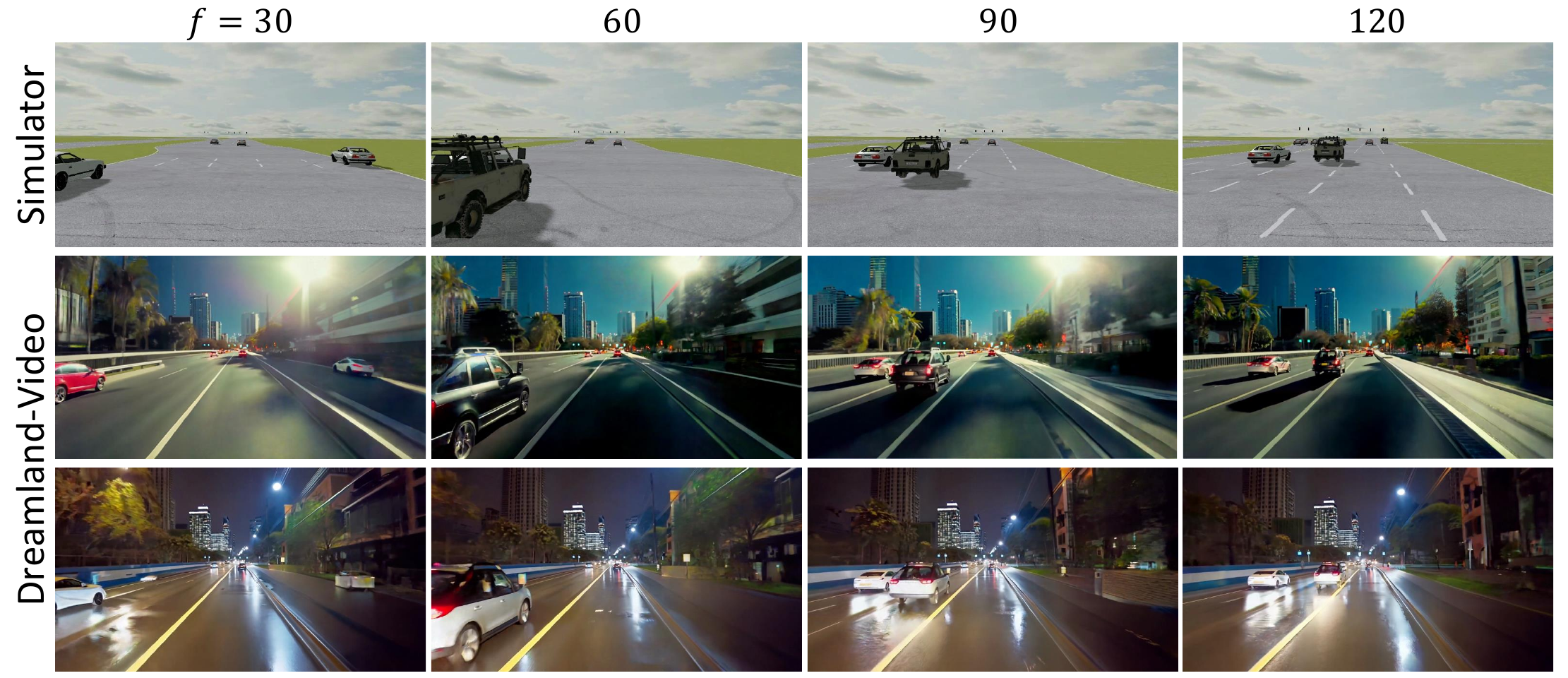}
    \vspace{-0.75em}
    \caption{
    \textbf{Qualitative Results}.
    Our \method-Video pipeline can re-render the simulator reference video into realistic and diverse video frames while maintaining the pretrained world knowledge.
    } 
    \vspace{-1.0em}

    \label{fig:qual video}
\end{figure}

For Stage-3, we employ a multi-condition video diffusion model, Cosmos-Transfer1-7B~\cite{nvidia2025cosmos}, without finetuning. Leveraging this state-of-the-art model, \method-Video can produce videos from 1K up to 4K resolution and as long as 121 frames. We evaluate our pipeline using 800 videos from the D3Sim dataset and report FID using all extracted frames. 
Table \ref{tab:video} and Figure \ref{fig:qual video} demonstrate the strong performance of our pipeline. By preserving the world knowledge embedded in these models, \method-Video achieves the same level of controllability as \method while consistently delivering high visual fidelity. 

We provide additional video results on our project webpage, which includes three sections: (1) Simulator-controlled scene generation, (2) Diverse text-controlled scene generation with the simulator, and (3) Simulator-controlled safety-critical driving scene generation \method-Video demonstrates strong performance in world creation with a physical simulator and generative models.
Its zero-shot capability to generate safety-critical scenarios allows efficient sampling of dangerous driving behaviors, laying the foundation for training autonomous driving agents with safe behavioral responses. It is worth noting that, by swapping in other pretrained world models, the \method-Video framework can be extended to even longer or fully autoregressive video re-rendering,  we leave further exploration of this promising direction to future research.

\begin{table}[]
\vspace{-0.5em}
\centering
\resizebox{0.75\linewidth}{!}{
\begin{tabular}{lcccccc}
\toprule[1.5pt]
                & \multicolumn{4}{c}{Visual Quality}                         & \multicolumn{2}{c}{Controllability} \\
\cmidrule(r){2-5} \cmidrule(r){6-7}
Method          &\textit{f-30} FID~$\downarrow$ & \textit{f-60} FID~$\downarrow$ &\textit{f-90} FID~$\downarrow$ & \textit{f-120} FID~$\downarrow$ & si-RMSE~$\downarrow$            & mIoU~$\uparrow$           \\
\midrule[0.5pt]
Dreamland-Video & 90.74        & 88.93        & 91.51        & 88.78         & 0.659	           & 0.717  \\
\bottomrule[1.5pt]
\end{tabular}
}
\vspace{-0.2em}
\caption{\textbf{Quantitative Results of \method-Video.} Our pipeline preserved strong video generation and outstanding condition-following capabilities while adapting to our world representations. }
\vspace{-1.5em}
\label{tab:video}
\end{table}

\subsection{Ablation Study}
\label{section: ablation}

We evaluate the effectiveness of our Stage-2 instructional editing model by comparing it against a baseline that directly generates the abstraction. Specifically, we finetune the Flux-Fill-Dev~\cite{flux2024} model via LoRA~\cite{hu2022lora}  for 4K iterations under two setting: 
(1) editing pipeline: our Real-LWA serves as targets, and we provide an editing mask to the model as input;
(2) generation pipeline: segmentation and depth maps of the realistic views derived from pretrained networks are set as the learning target.

\begin{figure}[]
    \centering
    \vspace{-1.0em}
    \includegraphics[width=0.98\linewidth]{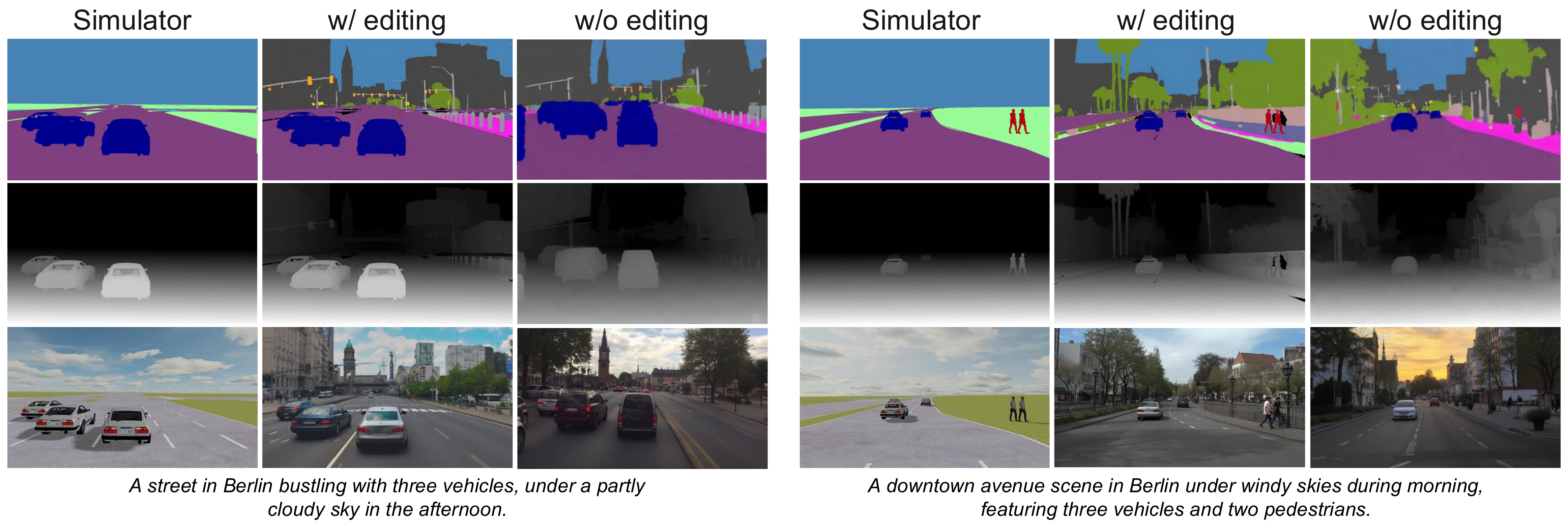}
    \vspace{-1em}
    \caption{
    \textbf{Qualitative Comparison} of Stage-2 model's design choices. 
    } 

    \label{fig: ablation stage1}
\end{figure}

At inference time, we again provide an editing mask to indicate regions that should be preserved to our editing models, and we employ the same Stage-3 model to render these condition maps into realistic scenes.  Across 16K generated samples on our validation set, qualitative results (Figure~\ref{fig: ablation stage1}) reveal that, without the editing approach, the Stage-2 model struggles to produce condition maps with sharp, accurate object boundaries (e.g., vehicles and pedestrians) and loses the fine‐grained control inherited from the simulator (e.g., specifying the exact height of a truck). We report the quantitative results in Table~\ref{tab:ablation} and additionally evaluate the refined condition maps' alignment to the simulator condition within the preserved region. Stage-2's editing model, trained with our editing data, constantly outperformed the baseline, showcasing the effectiveness of our pipeline design.

\begin{table}[]

\vspace{-1.25em}
\centering
\resizebox{0.75\linewidth}{!}{
\begin{tabular}{lccccc}
\toprule[1.5pt]
 & Image Quality & \multicolumn{2}{c}{Pipeline Controllability} & \multicolumn{2}{c}{Stage-2 Controllability} \\
 \cmidrule(r){2-2} \cmidrule(r){3-4} \cmidrule(r){5-6}
Method                & FID~$\downarrow$  & si-RMSE~$\downarrow$  & mIoU~$\uparrow$  & si-RMSE~$\downarrow$  & mIoU~$\uparrow$  \\
\midrule[0.5pt]
Dreamland w/o editing & 63.21 & 0.713 & 0.658 & 0.709 & 0.649 \\
Dreamland             & \textbf{50.58} & \textbf{0.647} &\textbf{ 0.791} & \textbf{0.672} & \textbf{0.950} \\
\bottomrule[1.5pt]

\end{tabular}
}
\vspace{-0.25em}
\caption{\textbf{Ablation Study on Stage-2 design choices.} Dreamland's Stage-2 design largely improved alignment to Stage-1's simulator, leading to better visual quality.}
\label{tab:ablation}
\end{table}

\subsection{Embodied Agent Training}
\label{section:agent training}
\begin{wrapfigure}{R}{0.55\textwidth}
    \centering
    \vspace{-1em}
    \begin{minipage}{0.55\textwidth}
    \begin{scriptsize}
    \resizebox{\linewidth}{!}{
        \begin{tabular}{lccc}
        \toprule
        Training Set         & Overall                     & Synthetic                       & Real \\
        \midrule
        MetaVQA-Synthetic       & 65.7\%                      & 65.7\%                    & 65.6\% \\
        MetaVQA+Dreamland    & \makecell{68.4\%\\(\textcolor{TealGreen}{+2.7\%})} & \makecell{67.3\%\\(\textcolor{TealGreen}{+1.6\%})} & \makecell{69.5\%\\(\textcolor{TealGreen}{+3.9\%})} \\
        \bottomrule
        \end{tabular}
    }
    \end{scriptsize}
    \captionof{table}{\textbf{Improved Model Learning with Dreamland}. InternVL2-8B, fine-tuned with Dreamland re-rendering, demonstrates significantly improved test accuracy on the curated MetaVQA test set.}
    \label{tab:metavqa}
    \end{minipage}
\end{wrapfigure}
We further demonstrate that the output synthetic data of Dreamland as data augmentation can enable the downstream task of embodied agent training. As presented in MetaVQA~\cite{wang2025metavqa}, synthetic images benefit the learning of situational awareness for general-purpose Vision-Language Models (VLMs) through Visual-Question-Answering. To validate improved learning from enriched visual observations through Dreamland, we set up two LoRA~\cite{hu2022lora} fine-tuning trials for InternVL2-8B~\cite{chen2024internvl}. Generated from an identical set of top-down layouts, the first training set's observation comprises only synthetic images rendered from the MetaDrive~\cite{li2021metadrive}. In comparison, half of the second training set's images are rendered via Dreamland. Question-answer pairs are generated for corresponding image sets, and the model is fine-tuned on the two equally-sized VQA sets and tested on a curated test set. Table~\ref{tab:metavqa} shows that the VLM model improves the test set especially on real-image-VQAs. These improvements showcase Dreamland's applicability in complex downstream tasks. Further experiment details will be in Appendix~\ref{sec:suppl_metavqa}.

\section{Conclusion}
\label{sec: conclusion}

We present Dreamland, a hybrid generation framework that, via a novel layered world abstraction (LWA), bridges physics-based simulators with large-scale pre-trained generative models to balance precise control and rich visual realism. To facilitate Sim2Real transfer for such systems, we introduce D3Sim, a large-scale dataset of paired simulated and real-world driving scenarios. Furthermore, we show  Dreamland's strong semantic and geometric control enhances the real-world adaptation of downstream embodied agents. 
We hope our pipeline design can shed light on Sim2Real generation and more broadly embodied agent learning with generative models. 

\textbf{Limitation}. \method introduced an additional editing model to the current hybrid pipeline, which led to a longer inference time. Also, the high-quality simulator and real-world paired data are expensive to annotate, which may limit the performance of the instructional editing model.

{
    \small
    \bibliographystyle{unsrt}

    \bibliography{main}
}

\newpage
\newpage
\newpage
\newpage

\setcounter{figure}{0}
\setcounter{table}{0}
\setcounter{section}{0}
\renewcommand{\thefigure}{\Alph{figure}}
\renewcommand{\thesection}{\Alph{section}}
\renewcommand{\thetable}{\Alph{table}}

\noindent\textbf{\large{Appendix}}

\appendix

\startcontents
{
    \hypersetup{linkcolor=black}
    \printcontents{}{1}{}
}
\newpage


\section{D3Sim Dataset}
\label{sec:suppl_data}
\subsection{D3Sim Dataset Construction}
\smallskip 
\noindent
\textbf{Data preparation}.

We construct our D3Sim dataset based on the nuPlan dataset \cite{nuplan}, which contains diverse real-world driving scenarios and more than 120 hours of sensor data. To obtain the driving data in the simulation domain, we reconstruct the digital twin of real-world driving scenarios in the MetaDrive simulator \cite{li2021metadrive} using ScenarioNet \cite{li2023scenarionet}. This results in scene records corresponding to more than 20,000 digital twin scenarios of 15-20 seconds in length and up to 10 Hz sample rate.

To get the conditions (\eg, depth, segmentation) in the simulation domain (\emph{Sim Conditions}), we replay the ego vehicle trajectory recorded in the scene record in the simulator, and capture the simulated sensor data from the front view camera. To achieve pixel-wise alignment between the digital twins, we calibrated the simulator sensors according to the camera parameters (intrinsics and extrinsics) recorded in the nuPlan dataset for each scenario. Then we capture the Sim Conditions synthesized using Panda3D engine based on OpenGL rendering backend, which includes depth map, semantic map, instance map, rendered RGB, and top-down view. The Sim-LWA is constructed from these Sim Conditions.

\smallskip 
\noindent
\textbf{Annotation}.
We annotate our digital twin scenarios to obtain realistic conditions (\emph{Real Conditions}) using pre-trained foundation models. Given a realistic front camera view from the nuPlan sensor data, we use DepthAnything2 \cite{depth_anything_v2} to obtain the depth map, SegFormer \cite{xie2021segformer} for the segmentation map, GroundingDino \cite{liu2023grounding} $+$ SAM2 \cite{ravi2024sam2segmentimages} for instance map, and Llama 3 \cite{grattafiori2024llama3herdmodels} for text descriptions. All of the Real Conditions align with the Sim Conditions in the pixel space, which enables us to construct pairwise training data for Sim2Real transfer described below.

\subsection{D3Sim Training}
\label{sec:d3sim_train}
\begin{figure}[h]
    \centering
    \vspace{-2em}
    \includegraphics[width=1.0\linewidth]{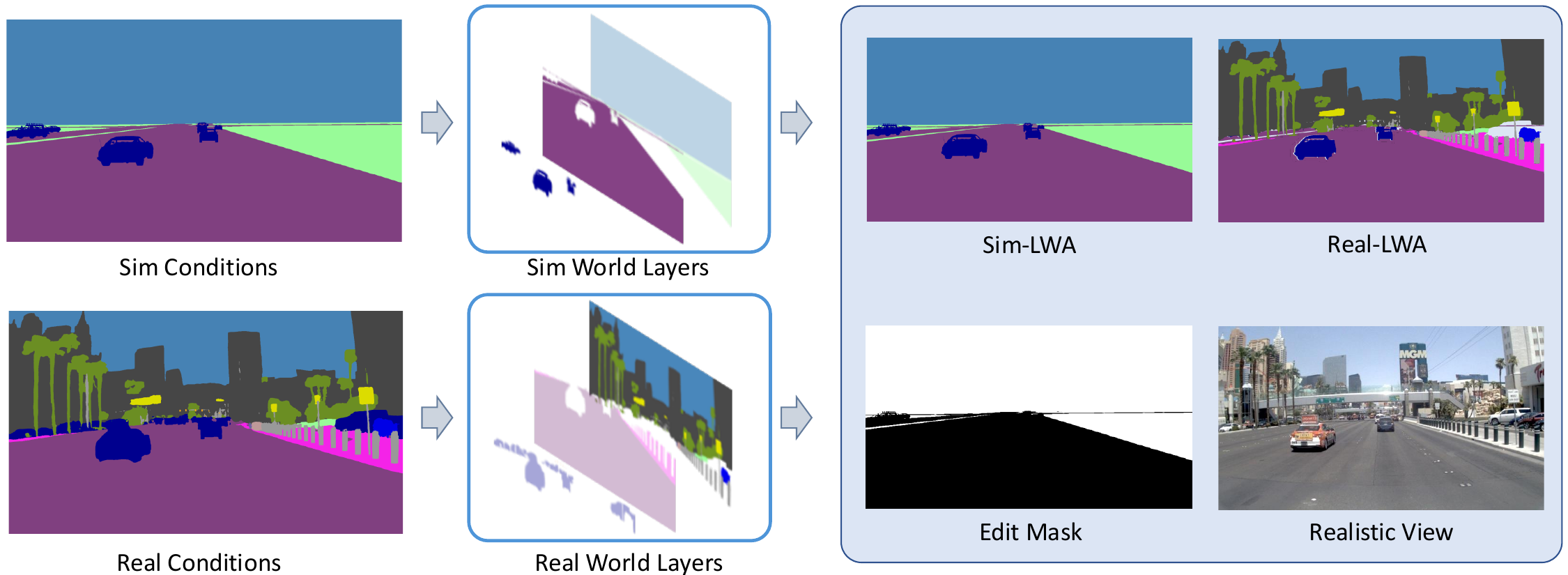}
    \caption{
    \textbf{Layered World Abstraction construction pipeline}. }
    \label{fig:edit_pipeline}
\end{figure}

Given the digital driving scenario with Sim and Real conditions, we construct our training dataset for the second and third stages. We first split the conditions into three world layers according to the segmentation mapping in Table~\ref{tab:layer_semantics}, the visibility mask $\mathcal{V}$ for each layer is derived as the pixels that belong to the corresponding semantic classes. Then we construct the concatenated conditions along the channel dimension, and construct the Sim-LWA and Real-LWA according to the predefined preserved and editable layers. The whole construction process is shown in Figure~\ref{fig:edit_pipeline}, where we only include the segmentation map as a condition for demonstration.

\begin{table}[h]
\centering
\resizebox{0.95\linewidth}{!}{
        \begin{tabular}{lc}
        \toprule
        World Layer         &  Semantic Classes \\
        \midrule
        Traffic Participants $\mathcal{L}^d$ & \colorbox{CAR}{\textcolor{white}{CAR}} \colorbox{TRUCK}{\textcolor{white}{TRUCK}} \colorbox{bus}{\textcolor{white}{BUS}} \colorbox{PEDESTRIAN}{\textcolor{white}{PEDESTRIAN}} \colorbox{bicycle}{\textcolor{white}{BICYCLE}} \colorbox{motorcycle}{\textcolor{white}{MOTORCYCLE}}  \\
        Map Layout $\mathcal{L}^l$ &  \colorbox{ROAD}{\textcolor{white}{ROAD}}  \colorbox{CROSSWALK}{\textcolor{white}{CROSSWALK}}  \colorbox{sidewalk}{\textcolor{white}{SIDEWALK}}  \colorbox{FENCE}{\textcolor{white}{FENCE}}  \colorbox{TRAFFICLIGHT}{\textcolor{white}{TRAFFIC\_LIGHT}} \colorbox{TRAFFICSIGN}{\textcolor{white}{TRAFFIC\_SIGN}} \\
        Background Layer $\mathcal{L}^b$ & \colorbox{SKY}{\textcolor{white}{SKY}} \colorbox{TERRAIN}{\textcolor{white}{TERRAIN}} \colorbox{BUILDING}{\textcolor{white}{BUILDING}} \colorbox{VEGETATION}{\textcolor{white}{VEGETATION}} \colorbox{WALL}{\textcolor{white}{WALL}} \colorbox{POLE}{\textcolor{white}{POLE}} \\
        \bottomrule
        \end{tabular}
    }
\caption{\textbf{Semantic mapping for world layers}.}
\label{tab:layer_semantics}
\end{table}

\begin{figure}[h]
    \centering
    \vspace{-2em}
    \includegraphics[width=1.0\linewidth]{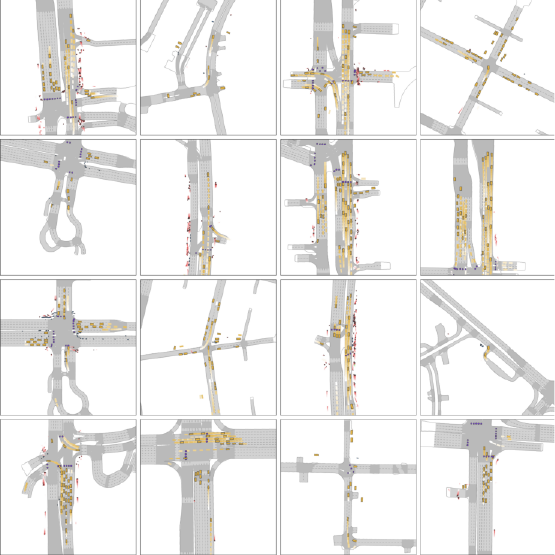}
    \caption{
    \textbf{Diverse Scene Layouts from D3Sim}. Ego vehicle and other traffic participants are marked with colored rectangles.}
    \label{fig:diverse_top_down}
\end{figure}

\smallskip 
\noindent
\textbf{Data Diversity}. 
To demonstrate the layout diversity of our D3Sim dataset, we visualize the top-down view of various driving scenarios in Figure~\ref{fig:diverse_top_down}, including unprotected cross turn, dense vehicle interactions, pickup/dropoff area, and following vehicle.

\subsection{D3Sim Validation}
We derive the D3Sim validation dataset from digital twins to evaluate controllability and visual quality. Despite driving scenarios from the nuPlan dataset reflecting real-world distribution, its complex object layout and interaction result in overlapping and occlusion, which hinders the accurate evaluation of controllability. Therefore, we construct an automated derivation pipeline to obtain scene records with diverse static layouts and unambiguous traffic participants' placement. 

Given a scene record, we first filter out occluded and non-visible objects according to the ego agent's view. Then we derived different variations of the scene by sampling from the combination of the objects. This process ensures that all objects in the sampled scenarios are clearly visible for controllability evaluation. We also randomly generate multiple diverse prompts for each sample, which are used as the text condition for the world model to evaluate its visual quality and diversity.

\subsection{D3Sim Video}
To extend our pipeline to video-based world models, we construct the video dataset with paired Sim and Real video conditions, along with realistic perspective view videos. The dataset contains around 20,000 video clips, each corresponding to a 15-20 second driving scenario captured at a 10 Hz sample rate. We follow the process described in Sec.~\ref{sec:d3sim_train} to construct the Sim-LWA, Real-LWA, and the edit mask for each frame, then concatenate the processed frames into videos. These paired videos are used to train the second stage of the \method-Video pipeline.

\section{Implementation Details}
\label{sec:suppl_imple_details}
\subsection{Dreamland Implementation Details}
We follow ACE++~\cite{mao2025ace++} to train our Stage-2 instructional editing model. Inspired by REVA~\cite{kara2024rave}, we vertically concatenate the depth and segmentation maps, each of size $512 \times 288$, into a single image of shape $512 \times 576$. We find that the instructional editing model can accurately generate structures in both the upper and lower conditional maps simultaneously, without any additional loss. We compare this approach with SimGen's method, which encodes the segmentation and depth maps into the Red and Blue channels, respectively. Empirically, we find that our approach performs better, as it is more aligned with the generative model's pre-training setting.

For the Stage-3 conditional generation model, we initialize it from the FLux Depth~\cite{flux2024} model and only train the linear projection layer $\xi$ to incorporate additional segmentation maps as conditioning inputs. To achieve this, we increase the input channels of the projection layer from 128 to 192, while keeping the output channel count unchanged. We also experimented with using LoRA to finetune the Stage-3 model; however, we empirically found that this approach tends to overfit to the finetuning dataset and degrade the model's world knowledge. Therefore, we only finetune the linear projection layer to preserve world knowledge while maintaining controllability.

\subsection{Dreamland-Video Implementation Details}
For the Stage-2 video-editing model, we increase the projection layer's input channel number by twice to take additional editing frames and editing masks conditions. Following standard practice in image editing models, we initialize from the pretrained generative model and concatenate the extra conditioning latents with the noise latent along the channel dimension.

\section{Experiments}
\subsection{Evaluation Metrics}

We evaluate \method from two key aspects: Generation Quality and Controllability. 

\smallskip
\noindent
\textbf{Generation Quality}.
For generation quality, we use Fréchet Inception Distance (FID)~\cite{heusel2017gans}. It measures the distribution distance of features between generated and original frames in the dataset. The features are extracted using a pre-trained Inception model. For quantitative results and comparisons, FID is evaluated on 10,000 samples from DIVA-Real~\cite{zhou2024simgen} to reflect the real-world driving distribution if not explicitly specified.

\smallskip
\noindent
\textbf{Controllability}.
For the quantitative evaluation of adherence to control input conditions, we transform the generated samples into shared representation spaces by applying depth estimation and semantic segmentation operations. We then compare the transformed representation with the original conditions captured from the simulator in the preserved region to measure the alignment between the generated world and the simulation. 

For depth alignment, we compute the scale-invariant Root Mean Squared Error (si-RMSE)~\cite{eigen2014depth} between the depth map from the simulation and transformed by DepthAnythingV2~\cite{depth_anything_v2}, where a lower value represents better alignment. For semantic alignment, we compute the mean Intersection over Union (mIoU) between the segmentation map from the simulation and the generated frames derived using SegFormer~\cite{xie2021segformer}. Higher value means better alignment.

\subsection{More Ablation}

\smallskip
\noindent
\textbf{Scalability}.
\method's simple pipeline design unlocks the potential of better image generation capability and controllability by upgrading the Stage-3 model to stronger image generative foundation models. Support by Table~{1}, Table~\ref{tab:supp baselines nus}, and user study results in Table~\ref{tab:user study}, \method could constantly combine with state-of-the-art conditional generation models to enhance the existing pipeline.

\smallskip
\noindent
\textbf{Real-LWA}. We ablate the effectiveness of our Stage-2 model design by forwarding Stage-3 with either \colorbox{green!15}{Sim-LWA} or \colorbox{blue!15}{Real-LWA}. As illustrated in Table~\ref{tab:ablation_lwa}, using Real-LWA consistently improves the FID, indicating that the generated content achieves better fidelity to real driving scenes. We also observe similar results in experiments on the nuScenes dataset, as recorded in Table~\ref{tab:supp baselines nus}. These observations highlight the necessity of using Real-LWA in a hybrid pipeline for scene creation.
\begin{table}[]
\resizebox{0.95\linewidth}{!}{
\begin{tabular}{lccccc}
\toprule[1.5pt]
\multirow{2}{*}{Methods}                & \multirow{2}{*}{Stage3 Input} & \multirow{2}{*}{Stage3 Model} & Image Quality & \multicolumn{2}{c}{Controllability} \\
\cmidrule{4-4}\cmidrule{5-6}
                                        &                               &                               & FID~$\downarrow$           & si-RMSE~$\downarrow$            & mIoU~$\uparrow$           \\
                                        \midrule
\multirow{7}{*}{Dreamland (Variations)} &  \cellcolor{green!15}Sim-LWA                       &  \cellcolor{yellow!15}Frozen SDXL                   & 106.80        & 0.806              & 0.505               \\
                                        &  \cellcolor{green!15}Sim-LWA                       & \cellcolor{orange!25}Frozen SD3                    & 78.30         & 0.808                 & 0.638               \\
                                        &  \cellcolor{green!15}Sim-LWA                       & \cellcolor{olive!35}Frozen Flux                   & 76.21         &    0.737              &   0.623             \\
                                        &  \cellcolor{green!15}Sim-LWA                       & \cellcolor{brown!50}Finetuned Flux                & 106.93        &  0.641                & 0.595
                                        \\
                                        \addlinespace
                                        \cmidrule{2-6}
                                        & \cellcolor{blue!15}Real-LWA                      &  \cellcolor{yellow!15}Frozen SDXL                   & 78.99         & 0.744          & 0.699         \\
                                        & \cellcolor{blue!15}Real-LWA                      & \cellcolor{orange!25}Frozen SD3                    & 51.65         & 0.673          & 0.747         \\
                                        & \cellcolor{blue!15}Real-LWA                      & \cellcolor{olive!35}Frozen Flux                   & 45.19         & 0.640         & 0.786         \\
Dreamland                               & \cellcolor{blue!15}Real-LWA                      & \cellcolor{brown!50}Finetuned Flux                & 44.61         & 0.647          & 0.791      \\
\bottomrule[1.5pt]
\end{tabular}
}
\caption{
\textbf{Ablation: Real-LWA} improves generation fidelity to real driving scene. 
}  
\label{tab:ablation_lwa}

\end{table}

\begin{figure}[]
    \centering
    \vspace{-2em}
    \includegraphics[width=1.0\linewidth]{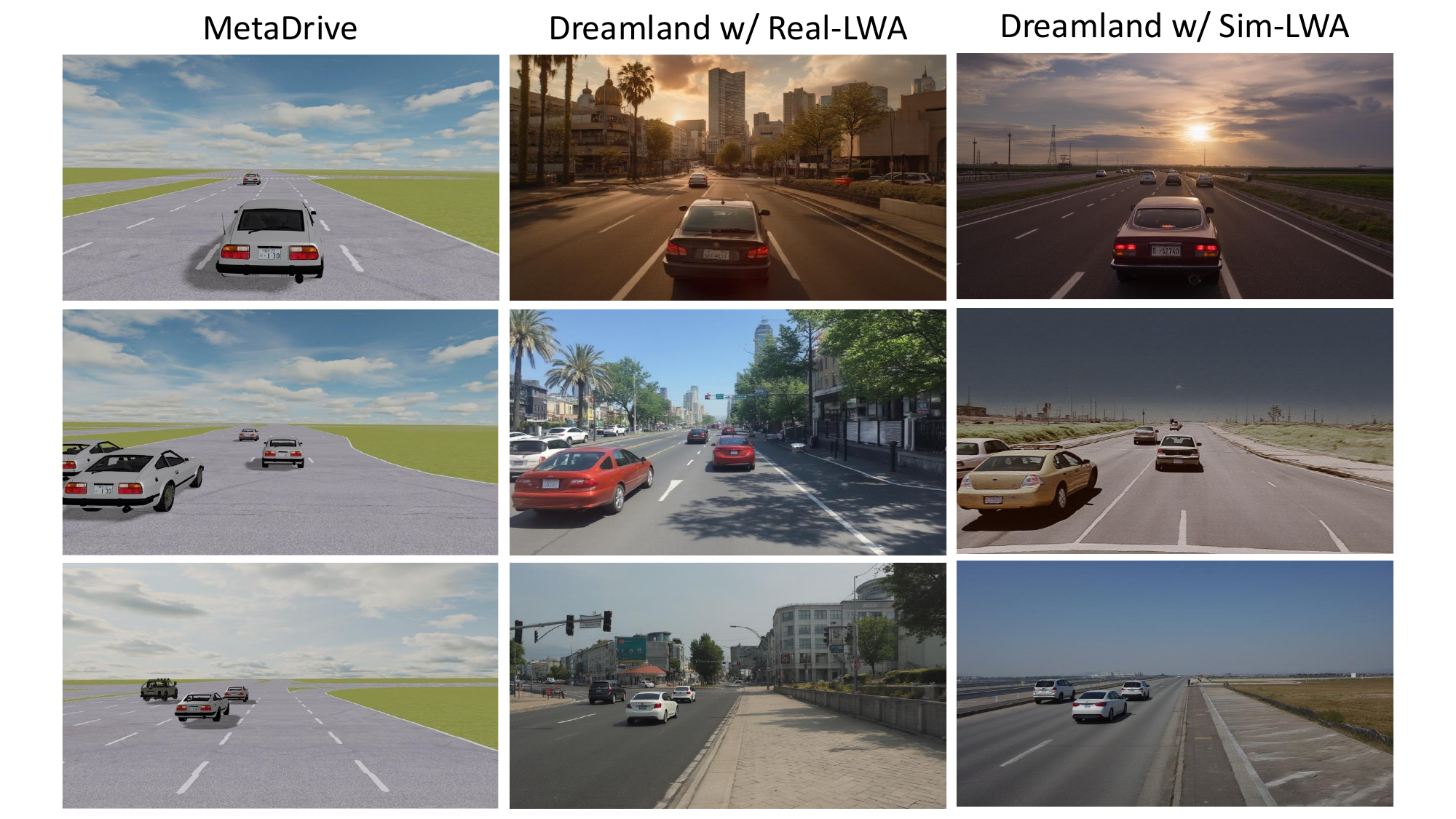}
    \caption{
\textbf{Ablation: Dreamland with Real-LWA and Sim-LWA}. Dreamland using Sim-LWA as the Stage-3 input generates driving scenes with depth identical to the MetaDrive conditions, but loses fidelity to real driving scenes.}

    \label{fig: supp ablation lwa}
\end{figure}

\subsection{Qualitative Results}
\begin{figure}[]
    \centering
    \vspace{-2em}
    \includegraphics[width=1.0\linewidth]{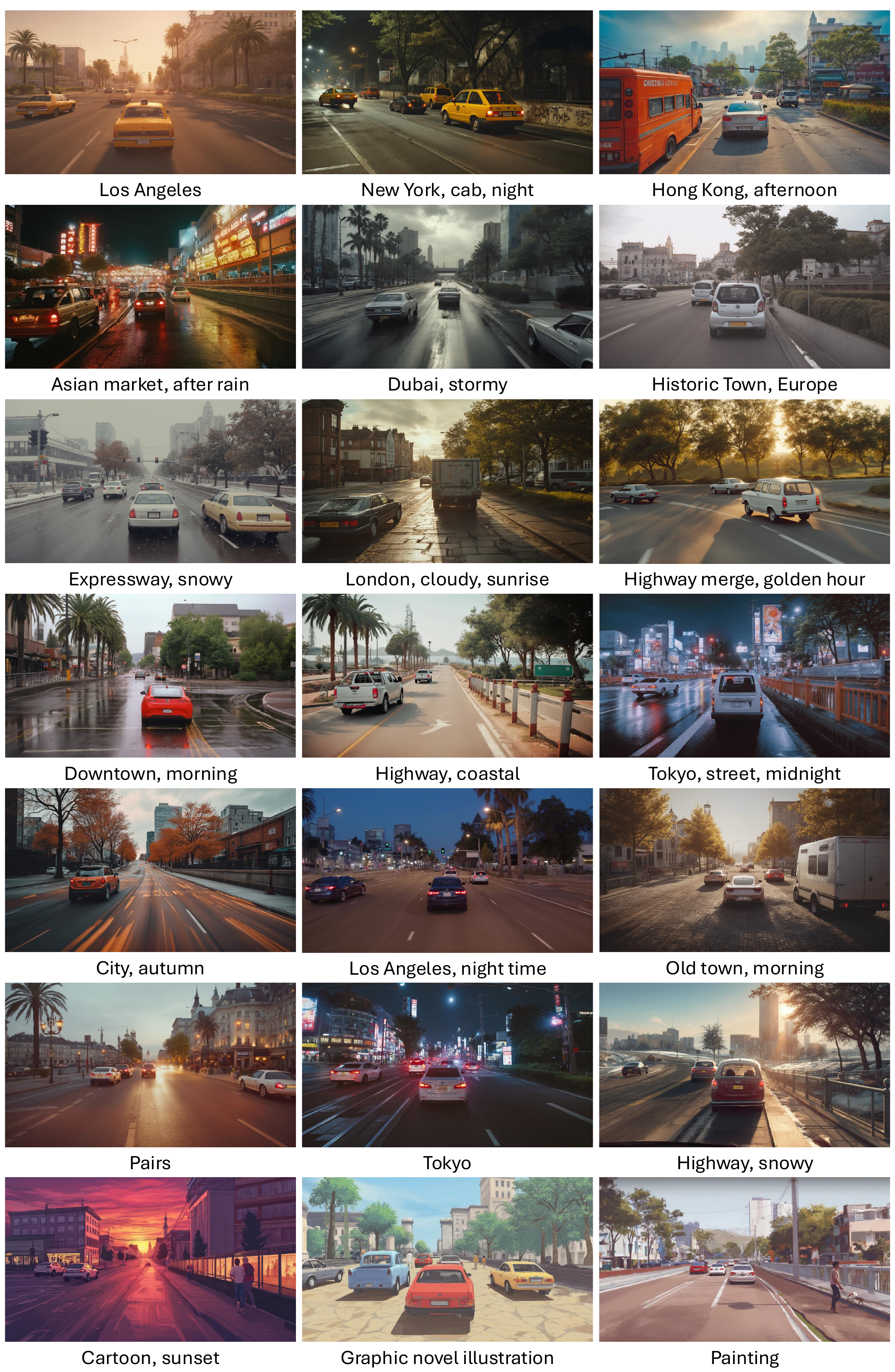}
    \caption{
    \textbf{Text-grounded scene creation with Dreamland}. \method could create diverse and visually realistic driving scenes by preserving the world knowledge in the pre-trained generative models. }

    \label{fig: supp text grounded generation}
\end{figure}

\smallskip
\noindent
\textbf{Alignment to text prompts}. Dreamland is capable of rendering the simulator into realistic frames while taking user-provided text prompts. As shown in Figure~\ref{fig: supp text grounded generation}, Dreamland can synthesize realistic scenes with different locations, weather, times, or style, demonstrating world knowledge preservation.

\begin{figure}[]
    \centering
    \vspace{-2em}
    \includegraphics[width=1.0\linewidth]{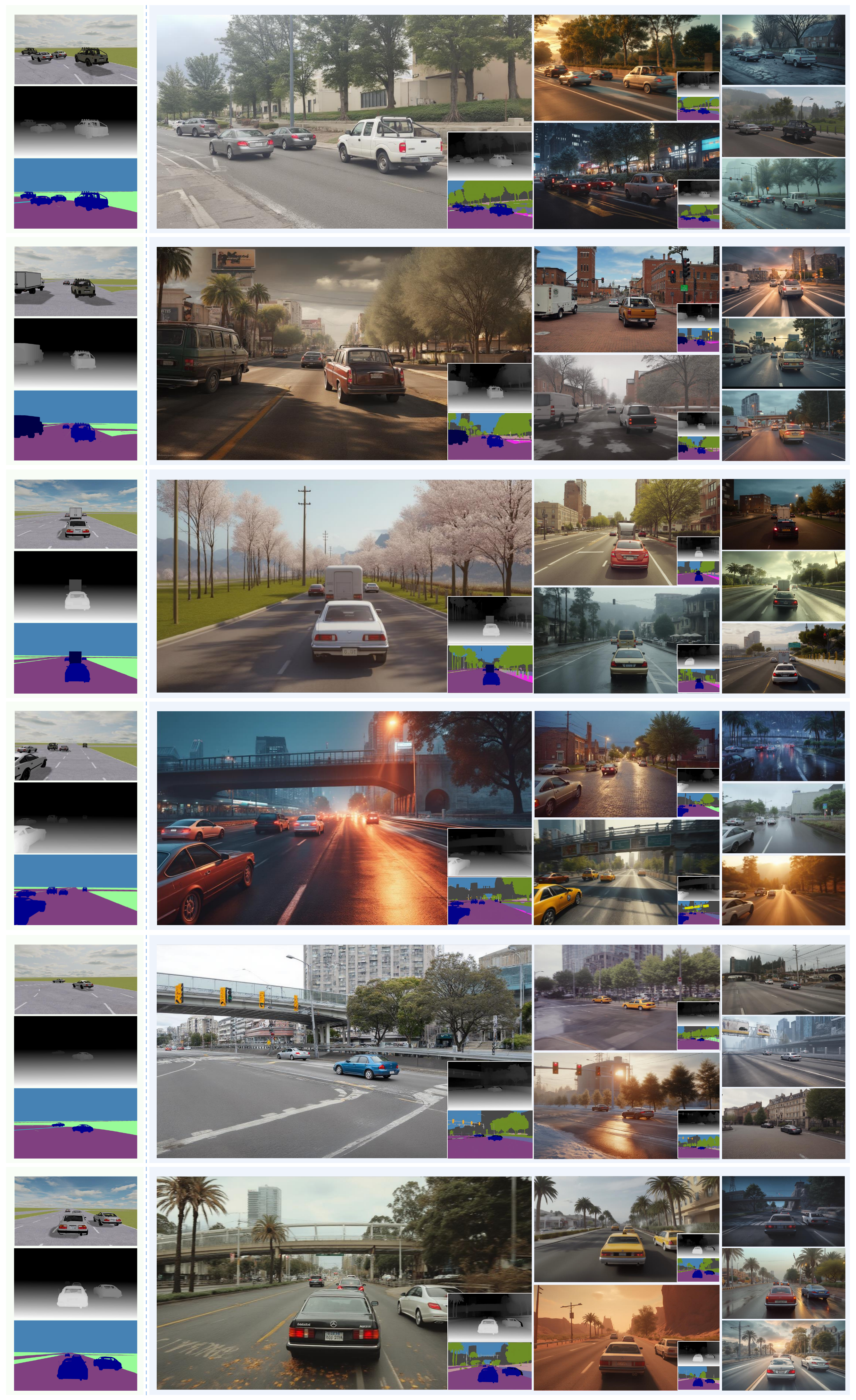}
    \vspace{-1em}
    \caption{
    \textbf{Rendering diverse appearances aligned to the simulator’s conditions.} \method can re-render complex driving scenes into realistic frames with preserved scene layout. }

    \label{fig: supp qual simulator generation}
\end{figure}

\smallskip
\noindent
\textbf{Alignment to simulator conditions}. We show more qualitative results of the \method in Figure~\ref{fig: supp qual simulator generation}. \method can re-render complex driving scenes into realistic frames with preserved scene layout.

\begin{figure}[]
    \centering
    \vspace{-2em}
    \includegraphics[width=1.0\linewidth]{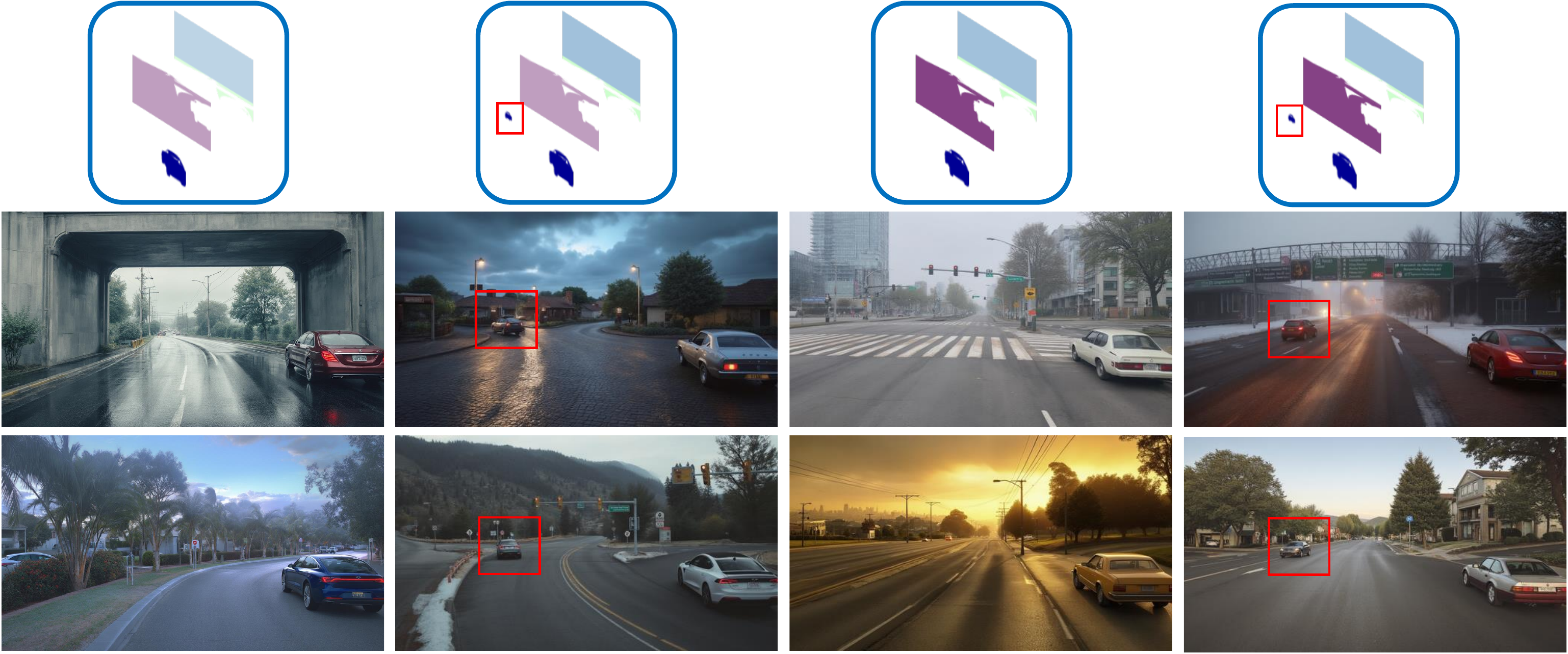}
    \vspace{-0.5em}
    \caption{
\textbf{Flexible control with LWA}. \textit{Top}: Visualization of Sim-LWA, where layers/objects not under user control are made transparent. \textit{Bottom}: Dreamland output. Users can choose to control specific vehicles or the road layout by adding them to or removing them from the preservation region. We use a red bounding box to highlight the small car that the user prefers to control in both the LWA and Dreamland outputs.
}

    \label{fig: qual lwa control}
\end{figure}

\smallskip
\noindent
\textbf{Flexible simulator control with LWA}. \method's LWA design can dynamically divide the simulator world into preserved and editable regions based on user instructions. This allows users to control specific vehicles or road layouts by adding them to or removing them from the preservation region. In Figure~\ref{fig: qual lwa control}, we show that \method's output closely aligns with the user-defined Sim-LWA and effectively respects the user-specified editable regions, indicating that fine-grained simulator control can be achieved through this hybrid approach.

\begin{figure}[h]
    \centering
    \includegraphics[width=1.0\linewidth]{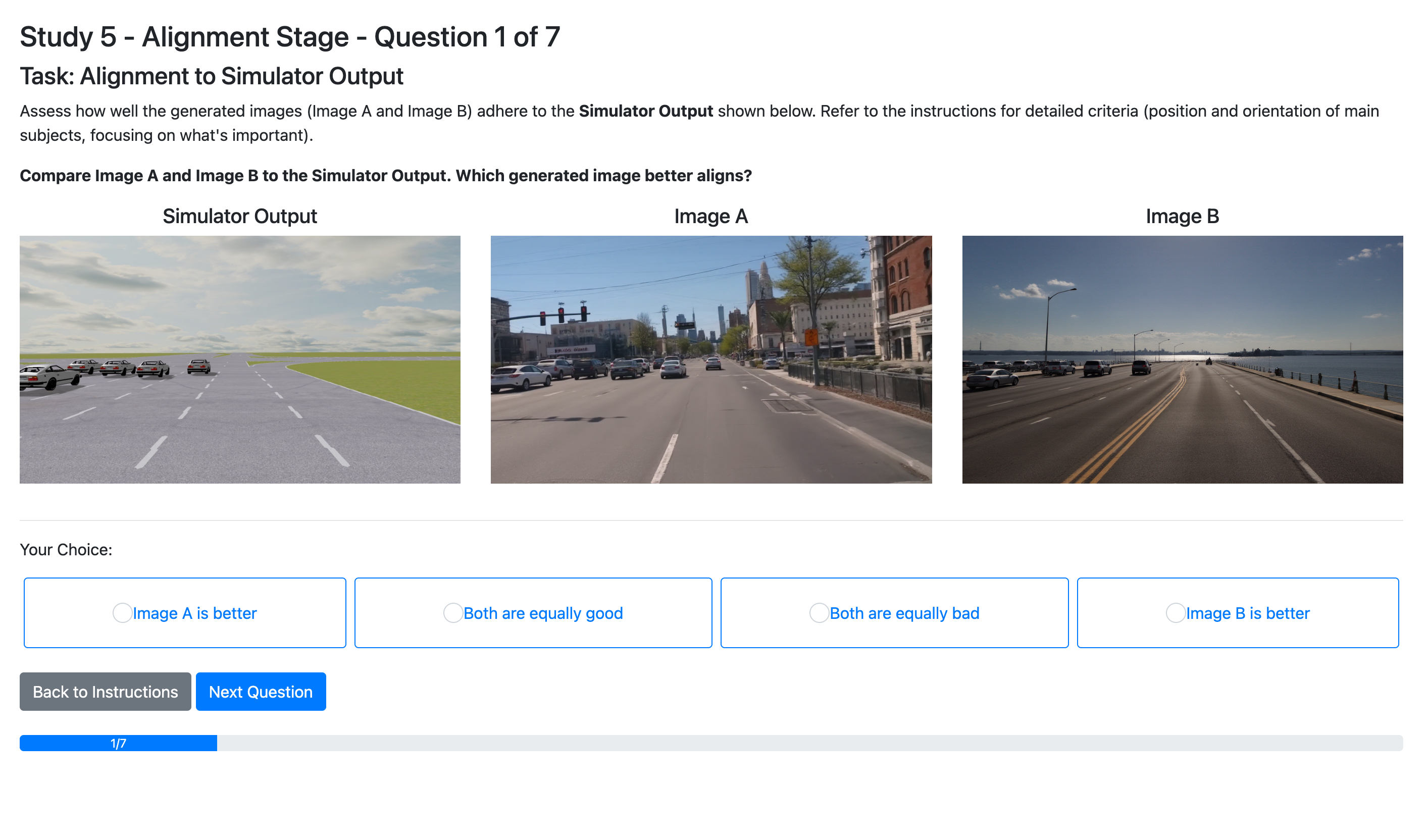}
    \vspace{-2em}
    \caption{
    \textbf{Screenshot of user study templates.} Users are asked to select between "Image A is better", "Image B is better", "Both are equally good" , and "Both are equally bad" options. We randomly shuffle the order of images in the user study.}

    \label{fig: user study}
\end{figure}

\smallskip
\noindent
\textbf{User study}.
To fully evaluate our method, we conducted a survey to study user preference between \method and its baseline methods and variations. We randomly selected 50 samples from our D3Sim validation dataset and compared the generated figures between ours and a baseline method regarding image quality and simulator controllability. Users are asked to choose from four options, including: "Image A is better", "Image B is better", "Both are equally good", and "Both are equally bad".  We recruited 25 users, each of whom evaluated 7 different samples, resulting in a total of 175 user votes for each comparison between our method and the baseline. Figure~\ref{fig: user study} shows our survey template and Table~\ref{tab:user study} reports the user preference between our methods and selected baselines.

In the comparison between Dreamland and Dreamland use Sim-LWA as the input for Stage-3, noted as Dreamland w/ Sim-LWA, using Real-LWA leads to 4.6 and 33.7 absolute percentage in image quality and controllability, showcasing the importance of our Stage-2 model.
It is also worth noting that \method largely outperforms previous state-of-the-art method (SimGen~\cite{zhou2024simgen}), achieving $95.4\%$ user preference in image quality and $71.4\%$ in simulator controllability.

\begin{table}[]
\centering
\resizebox{0.98\linewidth}{!}{
\begin{tabular}{lcccccccc}
\toprule[1.5pt]
            & \multicolumn{4}{c}{Image Quality}               & \multicolumn{4}{c}{Controllability}                    \\
            \cmidrule(r){2-5} \cmidrule(r){6-9}
Baseline              & Ours Better & Opponent Better & Both Good & Both Bad & Ours Better & Opponent Better & Both Good & Both Bad \\
\midrule
Dreamland w/ Sim-LWA     & \textbf{36.6 }     & 32.0       & 17.1    & 14.3    & \textbf{52.6}     & 18.9       & 24.6     & 4.0   \\
\addlinespace 

SimGen      & \textbf{95.4}      & 0.6       & 0.6     & 3.4    & \textbf{71.4}      & 10.3     & 6.3     & 12.0   \\
Dreamland~(\textit{Frozen SDXL}) & \textbf{90.9}      & 0.6       & 3.4     & 5.1    & \textbf{73.7}      & 9.1       & 10.3    & 6.9   \\
Dreamland~(\textit{Frozen SD3})  & \textbf{83.4}      & 4.6       & 6.3    & 5.7    & \textbf{86.9}     & 1.7       & 1.1     & 10.3   \\
Dreamland~(\textit{Frozen Flux}) & 33.7      & \textbf{44.0}      & 14.3    & 8.0    & \textbf{46.3}     & 31.4      & 12.6    & 9.7 \\
 \bottomrule[1.5pt]
\end{tabular}
}
\vspace{-0.5em}
\caption{\textbf{User Study}. We study the user preference by comparing \method and its baseline methods. The result suggests that \method achieved strong visual generation quality and outperforms all other baselines in alignment with simulator conditions.}
\label{tab:user study}
\end{table}

\subsection{Comparison with Other Baselines}

\smallskip
\noindent
\textbf{Experimental Setup.} We compare \method with previous generative models for driving scene generation, including BEVGen~\cite{bevgen}, BEVControl~\cite{yang2023bevcontrol}, MagicControl~\cite{gao2023magicdrive}, Panacea~\cite{wen2024panacea}, DrivingDiffusion~\cite{li2023drivingdiffusion}, and SimGen~\cite{zhou2024simgen}. SimGen shares the same setup as ours, employing a simulator to first render scene records and then using a generative model to re-render the simulator frames. The other baselines directly encode driving scene maps and generate driving scenes based on them. We conduct the experiment using the 6K spatial conditions from the nuScenes validation dataset and compute the FID with source images from nuScenes or DIVA-Real.

 \smallskip
\noindent
\textbf{nuScenes FID Assessment of Generative Models.} In recent years, it has been common practice for generative models in autonomous driving tasks to report FID scores based on natural driving images, such as nuScenes dataset~\cite{nuscenes2019}. However, recent large-scale foundation models~\cite{podell2023sdxl, esser2024sd3, flux2024} are often pre-trained on datasets rich in high-quality, visually aesthetic images. This leads them to generate stylized, artistic outputs that differ from naturally captured driving scenes. As a result, evaluating generative models for autonomous driving using nuScenes FID has become more and more questionable.
Zhou et al.~\cite{zhou2024simgen} observed that generative models trained on the DIVA-Real dataset demonstrate noticeably improved visual fidelity. Therefore, we compute FID using DIVA-Real, since it captures high-quality, aesthetically realistic driving scenes aligned with modern generative models.

While previous state-of-the-art models achieve low FID scores on nuScenes, their FID scores on DIVA-Real are significantly higher. For example, both Panacea~\cite{wen2024panacea} and SimGen~\cite{zhou2024simgen} achieve nuScenes FID scores around 16, but their FID scores on DIVA-Real exceed 60.
We validate the effectiveness of using FID on DIVA-Real through a human evaluation study, with results reported in Table~\ref{tab:user study}. In our user study, 95.4\% of participants agreed that \method produces higher visual quality compared to SimGen, and over 80\% preferred \method over the Dreamland variants using SDXL and SD3. This aligns with the FID scores on DIVA-Real, where \method significantly outperforms the aforementioned methods.
It is also worth noting that when \method and \method (\textit{Frozen Flux}) achieve similar FID scores (44.61 vs. 45.19), human preference is also consistent with the scores, with winning rates of 33.7\% and 44.0\%, respectively.

\smallskip
\noindent
\textbf{Quantitative results}. Table~\ref{tab:supp baselines nus} presents a comparison between \method and previous state-of-the-art models for autonomous driving tasks. Under the actual inference setting, \method achieves the lowest FID on DIVA-Real (44.61), outperforming the previous best method, MagicDrive, by 12.8\%, demonstrating the strong capability of \method in generating realistic driving scenes. Notably, \method using \colorbox{red!15}{ground truth condition maps} (e.g., depth and segmentation maps) achieves a similar FID score to \method with \colorbox{blue!15}{Real-LWA}, highlighting the effectiveness of our LWA design.

\begin{table}[]
\resizebox{1.0\linewidth}{!}{
\begin{tabular}{lcccc}
\toprule[1.5pt] 
                                  &               &                & \multicolumn{2}{c}{\textbf{Image Quality}} \\
                                        \cmidrule(r){4-5}
\textbf{Method}                                        &               &                & \textbf{FID - nuScenes}        & \textbf{FID - DIVA Real}      \\
                                        \midrule
BEVGen~\cite{swerdlow2024street}                                  &               &                & \color[HTML]{9b9b9b}25.50         & -                 \\
BEVControl~\cite{yang2023bevcontrol}                             &               &                & \color[HTML]{9b9b9b}24.90         & -                 \\
MagicDrive~\cite{gao2023magicdrive}                             &               &                & \color[HTML]{9b9b9b}16.60         & $51.20^{*}$             \\
Panacea~\cite{wen2024panacea}                                 &               &                & \color[HTML]{9b9b9b}16.96         & $61.83^{*}$             \\
DrivingDiffusion~\cite{li2023drivingdiffusion}                        &               &                & \color[HTML]{9b9b9b}15.90         & -                 \\
SimGen~\cite{zhou2024simgen}                                  &               &                & \color[HTML]{9b9b9b}15.60         & $68.20^{*}$            \\
\midrule
                                        & Stage3 Input  & Stage3 Model   &               &                   \\
\midrule
\multirow{11}{*}{Dreamland
(Variations)} & \cellcolor{red!15}GT Conditions & \cellcolor{yellow!15}Frozen SDXL    & \color[HTML]{9b9b9b}54.20         & 65.47             \\
                                        & \cellcolor{red!15}GT Conditions & \cellcolor{orange!25}Frozen SD3     & \color[HTML]{9b9b9b}43.68         & 50.01             \\
                                        &\cellcolor{red!15} GT Conditions & \cellcolor{olive!35}Frozen Flux    & \color[HTML]{9b9b9b}43.22         & 44.63             \\
                                        & \cellcolor{red!15}GT Conditions & \cellcolor{brown!50}Finetuned Flux & \color[HTML]{9b9b9b}42.36         & 42.27             \\
                                        \addlinespace
                                        \cmidrule(r){2-5}
                                        & \cellcolor{green!15}Sim-LWA      & \cellcolor{yellow!15}Frozen SDXL    & \color[HTML]{9b9b9b}93.66         & 100.57            \\
                                        & \cellcolor{green!15}Sim-LWA      & \cellcolor{orange!25}Frozen SD3     & \color[HTML]{9b9b9b}55.49         & 59.92             \\
                                        & \cellcolor{green!15}Sim-LWA      & \cellcolor{olive!35}Frozen Flux    & \color[HTML]{9b9b9b}52.66         & 51.68             \\
                                        & \cellcolor{green!15}Sim-LWA      & \cellcolor{brown!50}Finetuned Flux & \color[HTML]{9b9b9b}51.35         & 52.47             \\
                                         \addlinespace
                                         \cmidrule(r){2-5}
                                        &  \cellcolor{blue!15}Real-LWA      & \cellcolor{yellow!15}Frozen SDXL    & \color[HTML]{9b9b9b}68.69         & 78.99             \\
                                        &  \cellcolor{blue!15}Real-LWA      & \cellcolor{orange!25}Frozen SD3     & \color[HTML]{9b9b9b}48.73         & 51.65             \\
                                        &  \cellcolor{blue!15}Real-LWA      & \cellcolor{olive!35}Frozen Flux    & \color[HTML]{9b9b9b}46.00         & 45.19             \\
Dreamland                               &  \cellcolor{blue!15}Real-LWA      & \cellcolor{brown!50}Finetuned Flux & \color[HTML]{9b9b9b}47.93         & 44.61        \\

\bottomrule[1.5pt] 
\end{tabular}
}
\vspace{-0.5em}
\caption{ \textbf{Quantitative comparison with baseline methods}. All methods are evaluated using the spatial conditions from the nuScenes validation dataset, and the FID is computed with source images from nuScenes or DIVA-Real. For existing baselines, we use the reported FID on the nuScenes dataset from their respective publications, and compute the FID on DIVA-Real when open-source checkpoints are available. We mark scores reproduced by us with an asterisk (*).
}

\label{tab:supp baselines nus}

\end{table}

\subsection{Embodied Agent Training}
\label{sec:suppl_metavqa}
\begin{figure}[h]
    \centering
    \begin{subfigure}{0.49\linewidth}
        \captionsetup{justification=centering}
        \includegraphics[width=\textwidth]{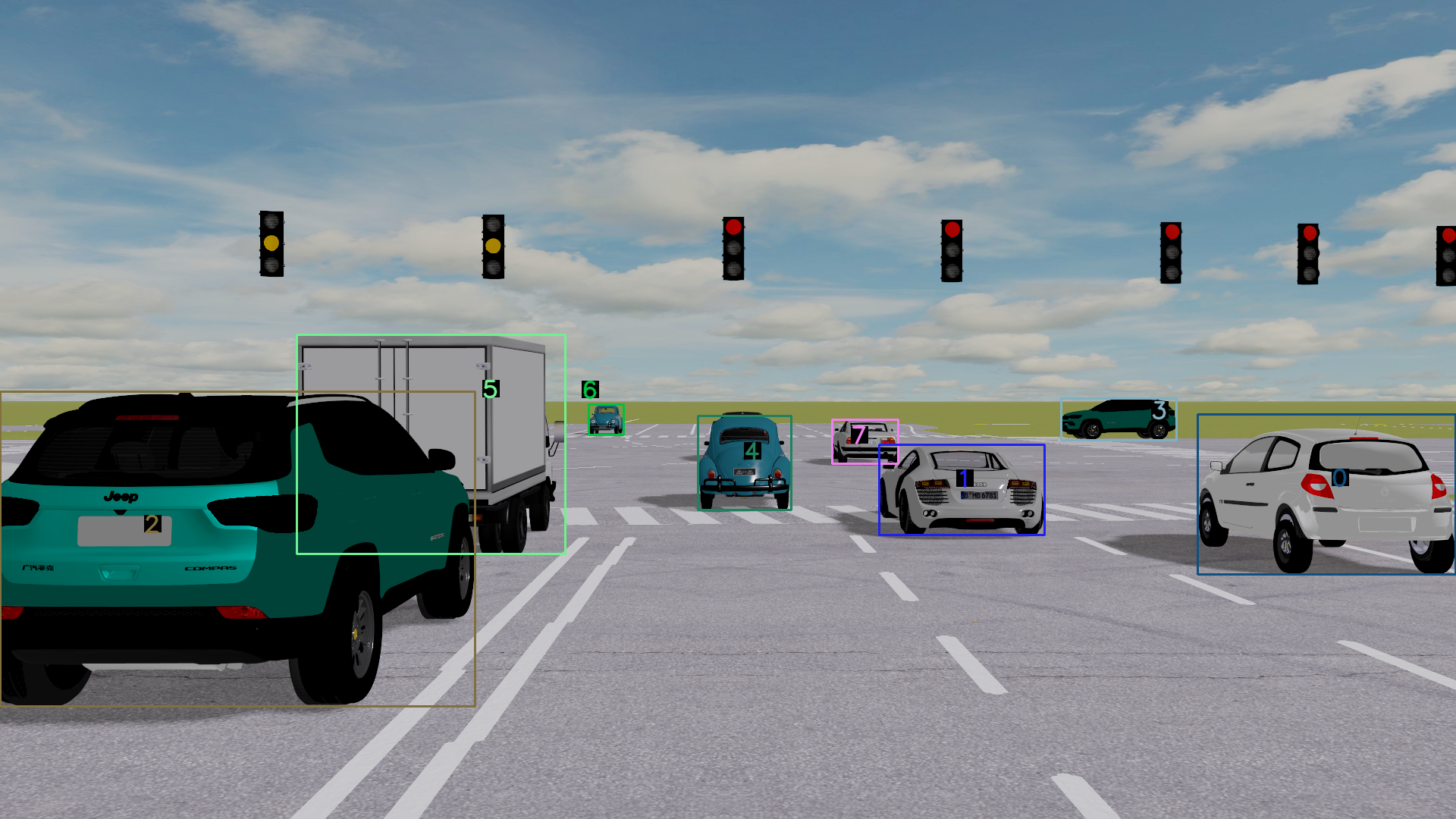}
        \caption*{\textbf{(a) MetaDrive-rendered Observation.}}
    \end{subfigure}
    \begin{subfigure}{0.49\linewidth}
        \captionsetup{justification=centering}
        \includegraphics[width=\textwidth]{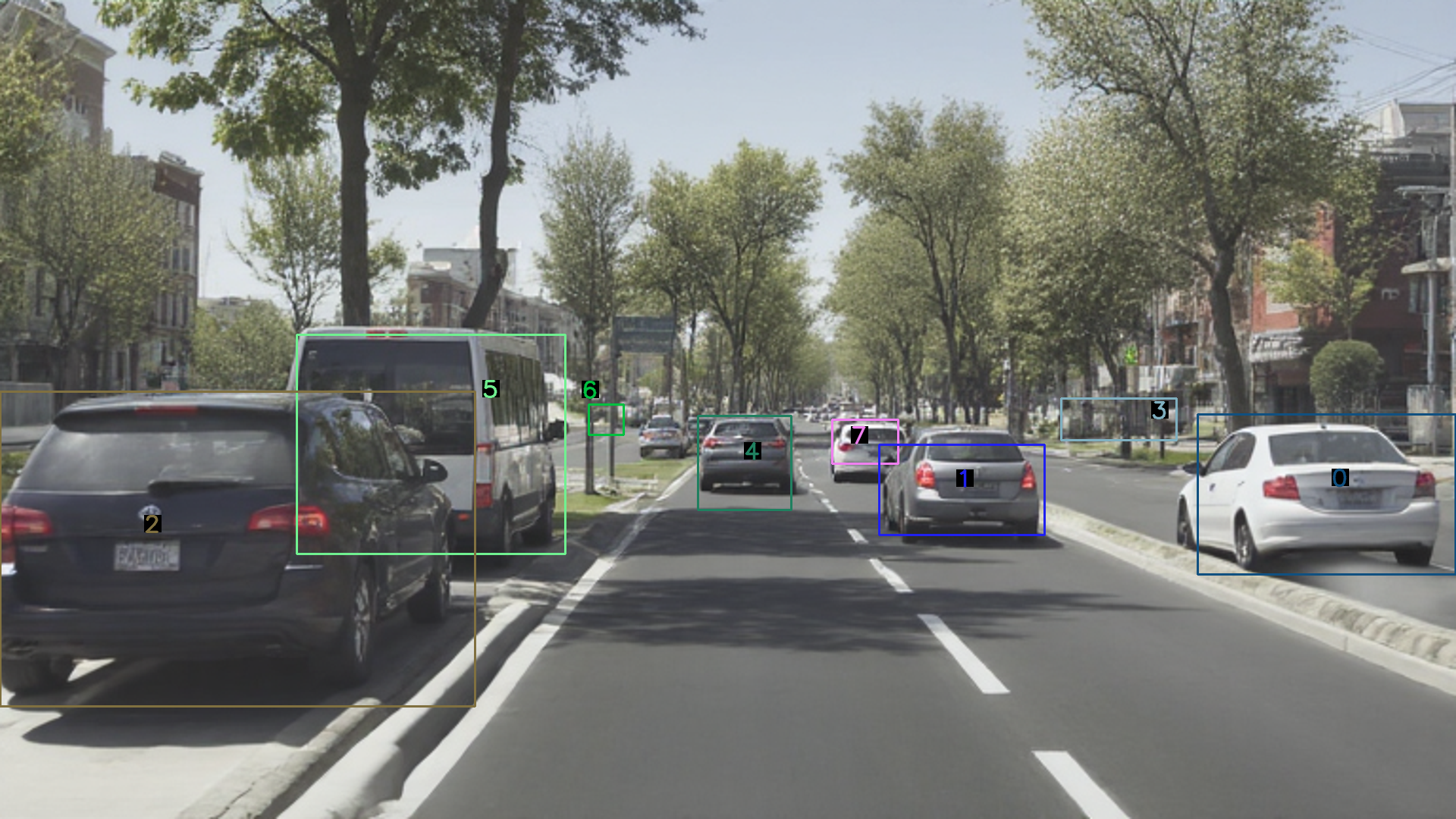}
        \caption*{\textbf{(b) Dreamland-rendered Observation.}}
    \end{subfigure}
    \caption{\textbf{Example questions}. \texttt{Suppose our current speed is slow(0-10 mph), and we perform action ``KEEP\_STRAIGHT" for 1.0 seconds. Will we run into object <2>, provided that it remains still? Select the best option from: (A) Yes; (B) No.} The answer is \texttt{(B)}}
    \label{fig: metavqa}
\end{figure}

We explain the experimental setting in greater detail in this section. To curate the two training sets for this task mentioned before, we use a shared set of traffic scenarios leveraging the Scenarionet~\cite{li2023scenarionet} data platform. These scenarios are rendered with both the MetaDrive~\cite{li2021metadrive} simulator and Dreamland(see Figure~\ref{fig: metavqa} for illustration). For each scenario, we generate questions using the MetaVQA~\cite{wang2025metavqa} annotation tools, and randomly sample 1000 VQA tuples for each observation domain. We use the InternVL2~\cite{chen2024internvl} codebase for the LoRA fine-tuning of pre-trained InternVL2-8 B. We train the model for two complete epochs and evaluate the model's performance on a test set, which is generated using nuScenes~\cite{nuscenes2019} and Waymo~\cite{Sun_2020_CVPR} datasets. The test set also comprises real images(from nuScenes) and simulator-rendered images(from Waymo) for holistic evaluation of the learned model under different visual domains. The answer parsing and metrics calculation directly use the toolkit provided by the MetaVQA codebase. Refer to the main paper for our results.

\end{document}